\documentclass[10pt,twocolumn,letterpaper]{article}
\pdfoutput=1
\usepackage{wacv}              

\usepackage{graphicx}
\usepackage{amsmath}
\usepackage{amssymb}
\usepackage{booktabs}

\usepackage{times}
\usepackage{epsfig}
\usepackage{tikz}
\usetikzlibrary{matrix}
\usepackage{subcaption}
\usepackage{caption}
\usepackage{pgfplots}
\usepgfplotslibrary{fillbetween}
\usepgfplotslibrary{colorbrewer}
    \pgfplotsset{
        compat=newest,
        cycle list/Set1-3,
        cycle multi list={
            mark list*\nextlist
            Set1-3\nextlist
        },
        label style={font=\LARGE},
        axis x line=bottom,
        axis y line=left,
        every tick label/.append style={font=\Large}
    }
\usepackage{pgfplotstable} 
\usepackage{tabularx}
\usepackage{multirow}
\usepackage{color,soul}
\newcommand*\rot[1]{\rotatebox{90}{#1}}

\usepackage{letterspace}
\newcommand{\ROMAN}{\textls*[-100]}

\usepackage{algorithm}
\usepackage{algpseudocode}

\usepackage[accsupp]{axessibility}  

\usepackage[pagebackref,breaklinks,colorlinks]{hyperref}

\usepackage[capitalize]{cleveref}
\crefname{section}{Sec.}{Secs.}
\Crefname{section}{Section}{Sections}
\Crefname{table}{Table}{Tables}
\crefname{table}{Tab.}{Tabs.}

\def\wacvPaperID{489} 
\def\confName{WACV}
\def\confYear{2024}

\begin{document}

\setlength{\abovedisplayskip}{7pt}
\setlength{\belowdisplayskip}{7pt}

\title{Bag of Tricks for Fully Test-Time Adaptation}


\author{Saypraseuth Mounsaveng$^{*}$, Florent Chiaroni, Malik Boudiaf, Marco Pedersoli, Ismail Ben Ayed\\
\tt\textit{ÉTS Montréal, Canada}}

\maketitle


\renewcommand{\thefootnote}{\fnsymbol{footnote}}
\footnotetext[1]{Corresponding author: saypraseuth.mounsaveng.1@etsmtl.net}
\footnotetext[2]{Code is available at \url{https://github.com/smounsav/tta_bot}}

\begin{abstract}
Fully Test-Time Adaptation (TTA), which aims at adapting models to data drifts, has recently attracted wide interest.
Numerous tricks and techniques have been proposed to ensure robust learning on arbitrary streams of unlabeled data. However, assessing the true impact of each individual technique and obtaining a fair comparison still constitutes a significant challenge.
To help consolidate the community's knowledge, we present a categorization of selected orthogonal TTA techniques, including small batch normalization, stream rebalancing, reliable sample selection, and network confidence calibration. We meticulously dissect the effect of each approach on different scenarios of interest.
Through our analysis, we shed light on trade-offs induced by those techniques between accuracy, the computational power required, and model complexity. We also uncover the synergy that arises when combining techniques and are able to establish new state-of-the-art results. 
\end{abstract}

\section{Introduction}

  \pgfplotstableread[row sep=\\,col sep=&]{
        bs & CIFAR10-C/ResNet26 & CIFAR10-C/ResNet26/STD & ImageNet-C/ResNet50-BN & ImageNet-C/ResNet50-BN/STD & ICR50BN+BoT & ICR50GN & ICR50GN+BoT & ICVitBaseLN & ICVitBaseLN+BoT\\
        256 &83.59733174641926 & 0.1 & 38.88493245 & 0.03330656314 & 42.17 & 31.17 & 36.96 & 51.22 & 58.76 \\
        128 &83.64999796549476 & 0.1 & 41.29719883 & 0.008488821901 & 44.35 & 30.84 & 40.08 & 51.15 & 59.22 \\
        64 &83.31066487630207 & 0.1 & 42.83346558 & 0.05835904713 & 46.57 & 28.77 & 43.52 & 51.07 & 59.56  \\
        32 &82.63799794514972 & 0.1 & 42.26591 & 0.1765731959 & 46.79 & 24.38 & 46.57 & 51.00 & 59.75  \\
        16 &79.92733205159502 & 0.1 & 39.42595461 & 0.1309639367 & 46.90 & 24.15 & 46.50 & 50.97 & 59.80  \\
        8 &73.83733139038081 & 0.1 & 33.30039919 & 0.03736369812 & 44.90 & 24.05 & 46.07 & 50.90 & 59.77  \\
        4 &68.63799819946286 & 0.1 & 20.81062171 & 0.08337440757 & 40.42 & 23.99 & 45.02 & 50.91 & 59.59  \\
        2 &31.978665669759064 & 0.1 & 5.530888753 & 0.00876388418 & 32.03 & 23.92 & 42.32 & 50.90 & 59.04  \\
        1 &10 & 0.1446222183 & 0.1 & 0.0025727916 & 20.31 & 23.90 & 39.70 & 50.89 & 54.68  \\
    }\resultsCIFAR

  \pgfplotstableread[row sep=\\,col sep=&]{
        bs & SAR/ResNet50-BN & Delta/ResNet50-BN & SAR/ResNet50-GN & Delta/ResNet50-GN & SAR/VitBase-LN & Delta/VitBase-LN \\
        256 & 39.59 & 40.48 & 32.97 & 35.76 & 53.53 & 57.77 \\
        128 & 42.21 & 42.80 & 34.33 & 38.94 & 53.85 & 58.34 \\
        64 & 44.02 & 45.27 & 37.53 & 42.31 & 56.76 & 58.71 \\
        32 & 44.51 & 46.87 & 39.18 & 45.27 & 56.87 & 58.90 \\
        16 & 41.03 & 46.33 & 39.32 & 45.22 & 56.87 & 58.95 \\
        8 & 31.10 & 43.67 & 38.80 & 44.70 & 56.92 & 58.86 \\
        4 & 18.90 & 39.16 & 37.61 & 43.47 & 56.69 & 58.57 \\
        2 & 6.78 & 31.26 & 35.66 & 40.77 & 55.71 & 57.98 \\
        1 & 0.14 & 20.25 & 33.86 & 23.91 & 53.16 & 50.89 \\        
      }\resultsoverviewothermethods
    
    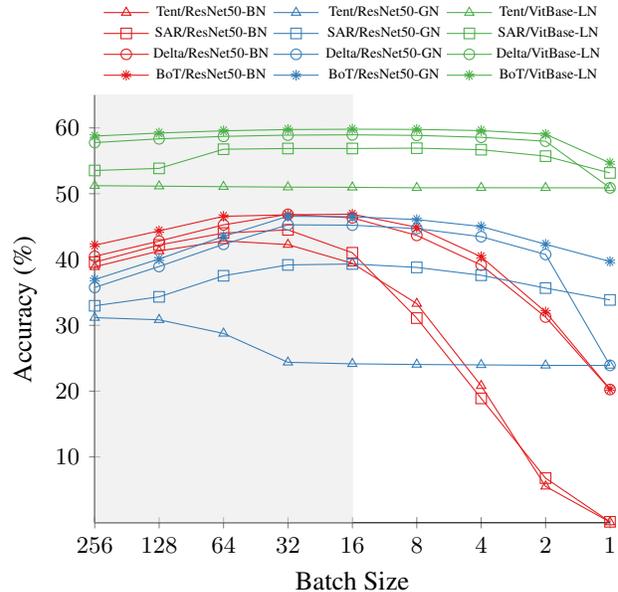
\begin{figure}[!tp]
        \centering
        \resizebox{\linewidth}{!}{%
            \begin{tikzpicture}
                \begin{axis}[
                    xlabel=Batch Size,
                    ylabel=Accuracy (\%),
                    axis line style={-},
                    label style={font=\normalfont},
                    xmin=1, xmax=256,
                    xmode=log,
                    log basis x=2,
                    log ticks with fixed point,
                    x dir=reverse,
                    ymin=0, ymax=65,
                    ytick={10,20,30,40,50,60,70,80,90,100},
                    ticklabel style={font=\small},
                    legend columns=4,
                    transpose legend,
                    legend style={
                                draw=none,
                                /tikz/column 3/.style={
                                 column sep=1pt,
                                },
                                font=\scriptsize,
                                at={(axis cs:256,80)},anchor=north west, nodes={scale=0.77, transform shape}},
                    cycle multi list={
                        Set1-3\nextlist
                        mark=triangle,mark=square,mark=o,mark=10-pointed star
                    },
                    ]
                    
                    \draw[fill=black!10, color=black!10, opacity=0.5] (axis cs:256,90) rectangle (axis cs:16,0);
                    
                    \addplot table[x=bs, y=ImageNet-C/ResNet50-BN]{\resultsCIFAR};
                    \addlegendentry{Tent/ResNet50-BN}
                    \addplot table[x=bs, y=SAR/ResNet50-BN]{\resultsoverviewothermethods};
                    \addlegendentry{SAR/ResNet50-BN}
                    \addplot table[x=bs, y=Delta/ResNet50-BN]{\resultsoverviewothermethods};
                    \addlegendentry{Delta/ResNet50-BN}  
                    \addplot table[x=bs, y=ICR50BN+BoT]{\resultsCIFAR};
                    \addlegendentry{BoT/ResNet50-BN}
                    \addplot table[x=bs, y=ICR50GN]{\resultsCIFAR};
                    \addlegendentry{Tent/ResNet50-GN}
                    \addplot table[x=bs, y=SAR/ResNet50-GN]{\resultsoverviewothermethods};
                    \addlegendentry{SAR/ResNet50-GN}
                    \addplot table[x=bs, y=Delta/ResNet50-GN]{\resultsoverviewothermethods};
                    \addlegendentry{Delta/ResNet50-GN}
                    \addplot table[x=bs, y=ICR50GN+BoT]{\resultsCIFAR};
                    \addlegendentry{BoT/ResNet50-GN}
                    \addplot table[x=bs, y=ICVitBaseLN]{\resultsCIFAR};
                    \addlegendentry{Tent/VitBase-LN}
                    \addplot table[x=bs, y=SAR/VitBase-LN]{\resultsoverviewothermethods};
                    \addlegendentry{SAR/VitBase-LN}
                    \addplot table[x=bs, y=Delta/VitBase-LN]{\resultsoverviewothermethods};
                    \addlegendentry{Delta/VitBase-LN}
                    \addplot table[x=bs, y=ICVitBaseLN+BoT]{\resultsCIFAR};
                    \addlegendentry{BoT/VitBase-LN}   
                \end{axis}
            \end{tikzpicture}
        }
        \caption{\textbf{Classification Accuracy in function of Batch Size for different methods and architectures on ImageNet-C.}  In this work, we choose to focus on small batches (16 and below, white zone). As the batch size decreases, the model performances remain stable until a batch size of 32 and then drops significantly for methods running on ResNet50-BN. Results reported are averaged over 15 corruptions and 3 runs. Confidence intervals are too small to be displayed.}
        \label{fig:intro}
    \end{figure}

    Deep neural networks perform well at inference time when test data comes from the same distribution as training data. However, they become inaccurate when there is a distribution shift \cite{QuioneroCandela2009DatasetSI}. This distribution shift can be caused by natural variations \cite{wilds2021} or corruptions \cite{hendrycks2019robustness, hendrycks2021many}. Test-Time Adaptation (TTA) aims at addressing this problem by adapting a model pre-trained on source data to make better predictions on shifted target data \cite{sun2020test, Iwasawa2021TestTimeCA, pmlr-v151-bartler22a}. In this work, we focus on the particular case of Fully Test-Time Adaptation (Fully TTA) \cite{Wang2021TentFT, niu2023towards, Zhao2023DELTADF}. In this setting, the adaptation is done source free and relies only on: i) a model pre-trained on data from a source domain and ii) unlabeled test data from a shifted target domain. Separating the training phase from the adaptation phase is particularly relevant for privacy-oriented applications where the training data is not available or can not be disclosed. 
    Fully TTA is also online. Test data is received as a continuous stream and the model adaptation is done on-the-fly as data is received. This makes the setup more realistic and closer to real-world "in-the-wild" scenarios where information about potential distribution shifts or about the quantity of data to be received is not necessarily available.

    Most of the recent solutions proposed to address Fully TTA are follow-ups of seminal work Tent \cite{Wang2021TentFT} and aim at solving problems inherent to the online and unsupervised aspect of Fully TTA. For example, \cite{Zhao2023DELTADF, Wang2016DealingWM} deal with the problem of the class imbalance data stream, \cite{niu2023towards, NEURIPS2022_fc28053a} improve the quality of the predictions used to adapt a model by selecting samples with a low entropy or leveraging the predictions of augmented samples and \cite{NEURIPS2022_fc28053a, Zhao2023DELTADF, Lim2023TTNAD, NEURIPS2022_fc28053a} investigate different normalization to stabilize the adaptation process.
    However, most of the tricks and techniques are presented in combination with others, which makes it difficult to identify their impact on the final model performance. Some techniques might already help when applied alone whereas others might only work or work better in combination with other tricks. 
    As this area of research is very active and developing fast, we aim in this study at disentangling the impact of some techniques recently proposed and evaluate objectively their contribution to the performance of Fully TTA models. We also propose possible improvements in specific cases. \\


    \textbf{Contribution.} To address the Fully Test-Time Adaptation problem, we analyzed the following techniques: i) Usage of batch renormalization or batch-agnostic normalization ii) Class re-balancing iii) Entropy-based sample selection iv) Temperature scaling. Those analyses were made considering small batch sizes (16 and below), which are closer to the potentially uncontrollable batch sizes of real-world scenarios.
    Our experimental results show that those techniques are already boosting the performance at test time when used alone, but that combining all of them leads to the best classification accuracy compared to a vanilla Tent method and 2 recent state-of-the-art methods on 4 different datasets. Additionally, to the accuracy improvement, the selected techniques also bring other interesting benefits like higher and more stable performance with small batch sizes and a reduced computational load by adapting the model with a reduced set of selected data.\\
    
    The remainder of the paper is structured as follows.  We conduct a literature review in Section \ref{sec:related_work}. Then we analyze each trick separately in a different section: architecture design in Sec.~\ref{sec:archnorm}, class rebalancing in Sec~\ref{sec:classrebalancing}, sample selection in Sec.~\ref{sec:sampleselection} and network calibration in Sec.~\ref{sec:calibration} before showing results on combinations of tricks in Sec.~\ref{sec:trickcombi} and results on other datasets in Sec.~\ref{sec:otherdatasets}. Finally, we conclude about the presented work in Sec.~\ref{sec:conclusion}.

\section{Related Work}
    \label{sec:related_work}

    \paragraph{Test-time adaptation (TTA).} Test-time adaptation assumes access to a pre-trained model and aims at leveraging unlabeled test instances from a (shifted) target distribution to make better predictions. Proposed methods usually employ one or a combination of the following techniques: \textit{self-training} to reinforce the model's own predictions through entropy minimization \cite{Wang2021TentFT} or Pseudo-Labelling schemes \cite{lee2013pseudo}, \textit{manifold regularization} to enforce smoother decision boundaries through data augmentation \cite{NEURIPS2022_fc28053a} or clustering \cite{Boudiaf2022ParameterfreeOT}, \textit{feature alignment} to mitigate covariate shift by batch norm statistic adaptation \cite{Li2016RevisitingBN, Schneider2020ImprovingRA}, and \textit{meta-learning} methods \cite{Goyal2022TestTimeAV} that try to meta-learn the best adaptation loss.
    
    \paragraph{TTA in the broader literature.} Although recently introduced \cite{Wang2021TentFT}, TTA shares important motivations and similarities with earlier or concurrent settings that are source-free domain adaptation (SFDA) \cite{Liang2020DoWR, Yang2021ExploitingTI, boudiaf2023search} and test-time training (TTT) \cite{sun2020test, Osowiechi2022TTTFlowUT}. In SFDA, methods also leverage samples from the target distribution of interest but have no access to source data, and the evaluation is still done on held-out test data. In other words, TTA is the transductive counterpart of SFDA. On the other hand, TTT works by constructing an auxiliary task that can be solved both at training and adaptation time and therefore, unlike TTA, is not agnostic to the training procedure or to the model architecture.

    \paragraph{Fully TTA.} TTA is of particular interest for online applications, in which the model receives samples as a stream. Operational requirements for online applications break crucial properties of the vanilla TTA setting e.g. large batch size or class balance. Under such operational requirements, standard TTA methods degrade, underperforming the non-adapted baseline and even degenerating to random performance in some cases \cite{Boudiaf2022ParameterfreeOT, niu2023towards}. Multiple regularization procedures have been proposed to address such shortcomings. Among them, (i) Improved feature alignment procedures that interpolate, between source and target statistics \cite{Nado2020EvaluatingPB, Lim2023TTNAD, Zhao2023DELTADF}, thereby improving overall estimation and decreasing reliance upon specific test batches, (ii) Sample re-weighting \cite{Zhao2023DELTADF, Niu2022EfficientTM} to alleviate the influence of class biases, (iii) Improving loss' intrinsic robustness to noisy samples, either encouraging convergence towards local minima \cite{niu2023towards} or preventing large deviations from the base model's predictions \cite{Boudiaf2022ParameterfreeOT, Niu2022EfficientTM}. Recently, \cite{Tang_2023_CVPR} explored the update of the model weights using Hebbian learning instead of just updating the BatchNorm layers. As this line of work grows, the current study provides an objective evaluation of how recently proposed ingredients translate into actual robustness for Fully TTA and quantifies the progress made so far, as well as pinpoints possible areas of improvement. A detailed comparison of the Fully TTA setting with the other TTA settings is available in the supplementary material.

\section{Experimental Setup}

    In this section, we present the details of our experimental setup. Firstly, we introduce the datasets used, then the different methods we want to compare and the different models, and finally, we explain the evaluation metric and protocol.
    For reproducibility purposes, the links to the code and model weights used in our experiments are provided in the supplementary material.

    \subsection{Datasets}
    We evaluate the different methods on several datasets used by prior SFDA or TTA studies:
    (i) ImageNet-C \cite{hendrycks2019robustness} is a variant of ImageNet \cite{ILSVRC15} where 19 corruption types and 5 levels of severity were applied. For our experiments, we report results using 15 corruption types at the most severe level of corruption (level 5) and keep the 4 remaining extra (speckle noise, gaussian blur, spatter, and saturate) as "validation" corruptions to select hyperparameters following \cite{Zhao2023DELTADF} and \cite{niu2023towards}. (ii) ImageNet-Rendition \cite{hendrycks2021many} consists of 30,000 images distributed in 200 Imagenet classes obtained by the rendition of ImageNet images like art, cartoons, tattoos, or video games. (iv) ImageNet-Sketch \cite{wang2019learning} is a dataset of 50,0000 images distributed in all ImageNet classes and obtained by querying Google Images with "sketch of \_\_" where \_\_ is the name of original ImageNet classes. Images are in the black and white color scheme. (v) Finally,  VisDA2017 \cite{visda2017} is a dataset of over 72K images distributed in 12 ImageNet classes and containing a mix of synthetic and real domain images.
    In the sections where we analyze tricks (Class rebalancing Sec.~\ref{sec:classrebalancing}, Sample Selection Sec.~\ref{sec:sampleselection}, Calibration Sec.~\ref{sec:calibration}, and Tricks combination Sec.~\ref{sec:trickcombi}), all experiments are done using ImageNet-C.

    \subsection{Methods}
    In this work, we chose to analyze the following tricks and methods: (i) Tent \cite{Wang2021TentFT} is a seminal work in Fully Test-Time Adaptation and is the first work to use an entropy-based loss in the adaptation process. (ii) SAR \cite{niu2023towards} is a state-of-the-art method in Fully TTA and proposes a method to select the most useful samples based on their entropy. (iii) Delta \cite{Zhao2023DELTADF} is also a state-of-the-art method in Fully TTA and focuses on addressing the problem of online class rebalancing. (iv) in our experimental setup, we call BoT the model combining the best tricks selected in the different experiments.

    \subsection{Models}
    In our experiments, we use different architectures depending on the datasets tested. In experiments with ImageNet-C, we follow \cite{niu2023towards} and use two variants of the ResNet50 architecture \cite{He2015DeepRL} and a ViT-Base/16 transformer architecture. The first ResNet50 variant (ResNet50-BN) uses batch normalization layers \cite{Ioffe2015BatchNA} whereas the second one (ResNet50-GN) uses group normalization \cite{Wu2018GroupN} layers. The ViTBase/16 transformer uses layer normalization \cite{Ba2016LayerN} and will be referred to as VitBase-LN. For experiments with VisDA2017, we follow \cite{Yang2021ExploitingTI} and \cite{boudiaf2023search} and use a ResNet101 architecture.
    The number of parameters of each architecture is available in the supplementary material.

    
    \subsection{Evaluation metrics}
    To evaluate the different approaches, we use the classification accuracy metric. To compute this metric, we follow \cite{niu2023towards} and \cite{Zhao2023DELTADF} and consider the accumulated predictions of the test samples after each model update. In other words, we do not compute the classification accuracy on the whole test set after the model has seen all test samples but online after each batch. Results reported are averaged over 3 runs.

\section{Architecture and Normalization}
    \label{sec:archnorm}
    In this section, we investigate the influence of different architectures and normalization on the model performance. Normalization in particular has been an active area of research in the TTA literature. \cite{Zhao2023DELTADF} shows that in the case of a distribution shift, normalization statistics are inaccurate within test mini-batches and the gradient of the loss can show strong fluctuations potentially destructive for the model. To address this issue, \cite{Lim2023TTNAD} proposes to combine linearly the statistics learned during training with the statistics computed at test time to reduce the gap between the source domain and the target domain. However, this method is not applicable in Fully TTA as it requires access to labeled source data to learn the linear combination in a post-training phase before using it at test time. \cite{NEURIPS2022_fc28053a, Mirza2021TheNM} also use a linear combination of the training statistics and the test statistics to handle the distribution shift. \cite{Zhao2023DELTADF} adapts batch renormalization \cite{Ioffe2017BatchRT} to test-time adaptation. Batch normalization parameters are updated using a combination of the mini-batch statistics and moving averages of these statistics like in the original paper, but in the TTA context, statistics and moving averages are computed using test batches. Another way to address the issues inherent to batch normalization is to use group or layer normalization instead as investigated in \cite{niu2023towards}. As the normalization differs a lot between works, this study aims at disentangling its effect from other techniques used.
    
    In our experiments, we follow \cite{niu2023towards} and use the following architectures: i) a ResNet50 with BatchNorm layers (ResNet50-BN) ii) a ResNet50 with GroupNorm layers (ResNet50-GN) iii) a VitBase/16 with LayerNorm layers (VitBase-LN) iv) to complete our pool of models to compare, we also include a variant of ResNet50-BN where batch normalization is replaced by batch renormalization (ResNet50-BReN).

\begin{figure}[!t]
        \centering
        \resizebox{0.5\columnwidth}{!}{%
            \begin{tikzpicture}
                \begin{axis}[
                    xlabel=Batch Size,
                    ylabel=Accuracy (\%),
                    axis line style={-},                    
                    label style={font=\Large},
                    xmin=1, xmax=16,
                    xmode=log,
                    log basis x=2,
                    log ticks with fixed point,
                    x dir=reverse,
                    ymin=0, ymax=55,
                    ytick={10,20,30,40,50,60,70,80,90,100},
                    cycle list/Set1-4,
                    cycle multiindex* list={
                        mark list*\nextlist
                        Set1-4\nextlist
                    },
                    legend columns=1, 
                    legend style={
                                overlay,
                                draw=none,
                                /tikz/column 2/.style={
                                 column sep=20pt,
                                },
                                at={(1.03,0.5)},
                                anchor=west,
                                nodes={scale=1.5, transform shape}
                                },
                    ]
                    \addplot table [x=bs, y=ResNet50-BN]{\resultsnormarchitecture};
                    \addplot table [x=bs,y=ResNet50-BReN]{\resultsnormarchitecture};
                    \addplot table [x=bs, y=ResNet50-GN]{\resultsnormarchitecture};
                    \addplot table[x=bs,y=VitBase-LN]{\resultsnormarchitecture};                    
                    \legend{ResNet50-BN, ResNet50-BReN, ResNet50-GN, VitBase-LN}
                \end{axis}
            \end{tikzpicture}
            \hspace*{0.1\textwidth}
        }
        \caption{\textbf{Impact of Normalization, Architecture, and Batch Size on classification accuracy of Tent method on ImageNet-C.} Using a batch renormalization layer leads to better performance than using a vanilla batch normalization. Tent performance is more stable on architectures with batch-agnostic normalization like group or layer normalization.}
        \label{fig:archinorm}            
    \end{figure}
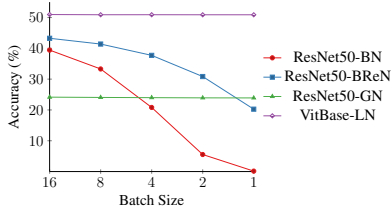

    \paragraph{Experimental results}
    In Fig.~\ref{fig:archinorm}, we observe that the performance of Tent method on a ResNet50-BN architecture is dropping when the batch size is becoming small, with a particularly low performance when the batch size is 2 ($5.53\%$ accuracy) or 1 ($0.14\%$ accuracy). Intuitively, those results can be explained by the fact that batch normalization layers are normalizing the weights based on the statistics of the current batch. When the batch becomes too small, the statistics computed have a high variance, are not representative anymore of the test distribution and are not informative enough about the domain shift.  However, we see that using batch renormalization instead of standard batch normalization improves the performance of a ResNet50 model and avoids a complete collapse of the model when the batch size is 1. Also in Fig.~\ref{fig:archinorm}, we observe that Tent performance on architectures with batch-agnostic normalization layers such as GroupNorm or LayerNorm is more stable and less impacted by a reduction of the batch size.
    
    \pgfplotstableread[row sep=\\,col sep=&]{
        bs & ResNet50-BN & ResNet50-BReN & ResNet50-GN & VitBase-LN \\
        16 & 39.42595461 & 43.26008784 & 24.15248823 & 50.96533203 \\
        8 & 33.30039919 & 41.38937668 & 24.05693284 & 50.8992431 \\
        4 & 20.81062171 & 37.71746574 & 23.98795491 & 50.90853221 \\
        2 & 5.530888753 & 30.83999953 & 23.92244371 & 50.8908435 \\
        1 & 0.1446222183 & 20.25173275 & 23.90155499 & 50.88835445 \\
    }\resultsnormarchitecture 
        
\section{Class rebalancing}
    \label{sec:classrebalancing}
    In this section, we explore the problem of online class imbalance in the context of Fully TTA. This problem is strongly relevant in this setting as data is received as a continuous stream. In this case, there is no guarantee that classes will appear in a balanced way or that different classes will appear in a given batch, especially when the batch size becomes much smaller than the total number of classes in the dataset. Imbalanced data can be particularly detrimental to the model performance as shown in \cite{Wang2016DealingWM,niu2023towards, Zhao2023DELTADF} and can lead in extreme cases to a model collapse to trivial solutions like assigning all samples to the dominant class.

\begin{figure}[!t]
        \centering
        \begin{subfigure}{0.33\columnwidth}
            \centering            
            \resizebox{\columnwidth}{!}{
                \begin{tikzpicture}
                    \begin{axis}[
                        xlabel=Imbalance Factor,
                        ylabel=Accuracy (\%),
                        xmin=1, xmax=6000,
                        xtick=data,
                        xtick={1, 1000 ,2000, 3000, 4000 ,5000, 6000},
                        xticklabels={1, 1K ,2K, 3K, 4K ,5K, $\infty$},
                        ymin=0, ymax=60,
                        ytick={10,20,30,40,50,60,70,80,90,100},
                        legend columns=3,                         
                        legend style={
                                    draw=none,
                                    /tikz/every even column/.style={
                                     column sep=5pt,
                                    },
                                    at={(axis cs:6000,69)},anchor=north east,nodes={scale=1.7, transform shape}}
                        ]
                        \addplot table [x=imbfactor, y=Tent]{\resultsclassrebalancingresnetbnsixteen};
                        \addplot table [x=imbfactor,y=DOT]{\resultsclassrebalancingresnetbnsixteen};
                        \addplot table [x=imbfactor, y=SAR]{\resultsclassrebalancingresnetbnsixteen};
                        \legend{Tent, DOT, SAR}
                    \end{axis}
                \end{tikzpicture}        
            }
            \caption{ResNet50-BN-16}
        \end{subfigure}%
        \hfill
        \begin{subfigure}{0.33\columnwidth}
            \resizebox{\columnwidth}{!}{
                \begin{tikzpicture}
                    \begin{axis}[
                        xlabel=Imbalance Factor,
                        ylabel=Accuracy (\%),
                        xmin=1, xmax=6000,
                        xtick=data,
                        xtick={1, 1000 ,2000, 3000, 4000 ,5000, 6000},
                        xticklabels={1, 1K ,2K, 3K, 4K ,5K, $\infty$},
                        ymin=10, ymax=50,
                        ytick={10,20,30,40,50,60,70,80,90,100},
                        legend columns=3,                         
                        legend style={
                                    draw=none,
                                    /tikz/every even column/.style={
                                     column sep=5pt,
                                    },
                                    at={(axis cs:6000,56)},anchor=north east,nodes={scale=1.7, transform shape}}
                        ]
                        \addplot table [x=imbfactor, y=Tent]{\resultsclassrebalancingresnetgnsixteen};
                        \addplot table [x=imbfactor,y=DOT]{\resultsclassrebalancingresnetgnsixteen};
                        \addplot table [x=imbfactor, y=SAR]{\resultsclassrebalancingresnetgnsixteen};
                        \legend{Tent, DOT, SAR}
                    \end{axis}
                \end{tikzpicture}
            }
                \caption{ResNet50-GN-16}
            \end{subfigure}%
            \hfill
            \begin{subfigure}{0.33\columnwidth}         
                \resizebox{\columnwidth}{!}{
                    \begin{tikzpicture}
                        \begin{axis}[
                            xlabel=Imbalance Factor,
                            ylabel=Accuracy (\%),
                            xmin=1, xmax=6000,
                            xtick=data,
                            xtick={1, 1000 ,2000, 3000, 4000 ,5000, 6000},
                            xticklabels={1, 1K ,2K, 3K, 4K ,5K, $\infty$},
                            ymin=40, ymax=65,
                            ytick={10,20,30,40,50,60,70,80,90,100},
                        legend columns=3,                         
                        legend style={
                                    draw=none,
                                    /tikz/every even column/.style={
                                     column sep=5pt,
                                    },
                                    at={(axis cs:6000,69)},anchor=north east,nodes={scale=1.7, transform shape}}
                            ]
                        \addplot table [x=imbfactor, y=Tent]{\resultsclassrebalancingvitbasesixteen};
                        \addplot table [x=imbfactor,y=DOT]{\resultsclassrebalancingvitbasesixteen};
                        \addplot table [x=imbfactor, y=SAR]{\resultsclassrebalancingvitbasesixteen};
                            \legend{Tent, DOT, SAR}
            
                        \end{axis}
                    \end{tikzpicture}
                }
                \caption{VitBase-LN-16}
            \end{subfigure}
            \medskip
            \begin{subfigure}{0.33\columnwidth}
                \centering            
                \resizebox{\columnwidth}{!}{
                    \begin{tikzpicture}
                        \begin{axis}[
                            xlabel=Imbalance Factor,
                            ylabel=Accuracy (\%),
                            xmin=1, xmax=6000,
                            xtick=data,
                            xtick={1, 1000 ,2000, 3000, 4000 ,5000, 6000},
                            xticklabels={1, 1K ,2K, 3K, 4K ,5K, $\infty$},
                            ymin=1, ymax=60,
                            ytick={10,20,30,40,50,60,70,80,90,100},
                            legend columns=3,                         
                            legend style={
                                        draw=none,
                                        /tikz/every even column/.style={
                                         column sep=5pt,
                                        },
                                        at={(axis cs:6000,69)},anchor=north east,nodes={scale=1.7, transform shape}}
                            ]
                        \addplot table [x=imbfactor, y=Tent]{\resultsclassrebalancingresnetbneight};
                        \addplot table [x=imbfactor,y=DOT]{\resultsclassrebalancingresnetbneight};
                        \addplot table [x=imbfactor, y=SAR]{\resultsclassrebalancingresnetbneight};
                            \legend{Tent, DOT, SAR}
            
                        \end{axis}
                    \end{tikzpicture}        
                }
                \caption{ResNet50-BN-8}
            \end{subfigure}%
            \hfill
            \begin{subfigure}{0.33\columnwidth}
                \resizebox{\columnwidth}{!}{
                    \begin{tikzpicture}
                        \begin{axis}[
                            xlabel=Imbalance Factor,
                            ylabel=Accuracy (\%),
                            xmin=1, xmax=6000,
                            xtick=data,
                            xtick={1, 1000 ,2000, 3000, 4000 ,5000, 6000},
                            xticklabels={1, 1K ,2K, 3K, 4K ,5K, $\infty$},
                            ymin=10, ymax=50,
                            ytick={10,20,30,40,50,60,70,80,90,100},
                        legend columns=3,                         
                        legend style={
                                    draw=none,
                                    /tikz/every even column/.style={
                                     column sep=5pt,
                                    },
                                    at={(axis cs:6000,56)},anchor=north east,nodes={scale=1.7, transform shape}}
                            ]
                        \addplot table [x=imbfactor, y=Tent]{\resultsclassrebalancingresnetgneight};
                        \addplot table [x=imbfactor,y=DOT]{\resultsclassrebalancingresnetgneight};
                        \addplot table [x=imbfactor, y=SAR]{\resultsclassrebalancingresnetgneight};
                            \legend{Tent, DOT, SAR}
            
                        \end{axis}
                    \end{tikzpicture}
                }
                    \caption{ResNet50-GN-8}
                \end{subfigure}%
                \hfill
                \begin{subfigure}{0.33\columnwidth}         
                    \resizebox{\columnwidth}{!}{
                        \begin{tikzpicture}
                            \begin{axis}[
                                xlabel=Imbalance Factor,
                                ylabel=Accuracy (\%),
                                xmin=1, xmax=6000,
                                xtick=data,
                                xtick={1, 1000 ,2000, 3000, 4000 ,5000, 6000},
                                xticklabels={1, 1K ,2K, 3K, 4K ,5K, $\infty$},
                                ymin=40, ymax=65,
                                ytick={10,20,30,40,50,60,70,80,90,100},
                            legend columns=3,                         
                            legend style={
                                        draw=none,
                                        /tikz/every even column/.style={
                                         column sep=5pt,
                                        },
                                        at={(axis cs:6000,69)},anchor=north east,nodes={scale=1.7, transform shape}}
                                ]
                                \addplot table [x=imbfactor, y=Tent]{\resultsclassrebalancingvitbaseeight};
                                \addplot table [x=imbfactor,y= DOT]{\resultsclassrebalancingvitbaseeight};
                                \addplot table [x=imbfactor, y=SAR]{\resultsclassrebalancingvitbaseeight};
                                \legend{Tent, DOT, SAR}
                
                            \end{axis}
                        \end{tikzpicture}
                    }
                    \caption{VitBase-LN-8}
                \end{subfigure}
            \medskip
            \begin{subfigure}{0.33\columnwidth}
                \centering            
                \resizebox{\columnwidth}{!}{
                    \begin{tikzpicture}
                        \begin{axis}[
                            xlabel=Imbalance Factor,
                            ylabel=Accuracy (\%),
                            xmin=1, xmax=6000,
                            xtick=data,
                            xtick={1, 1000 ,2000, 3000, 4000 ,5000, 6000},
                            xticklabels={1, 1K ,2K, 3K, 4K ,5K, $\infty$},
                            ymin=0, ymax=60,
                            ytick={10,20,30,40,50,60,70,80,90,100},
                            legend columns=3,                         
                            legend style={
                                        draw=none,
                                        /tikz/every even column/.style={
                                         column sep=5pt,
                                        },
                                        at={(axis cs:6000,69)},anchor=north east,nodes={scale=1.7, transform shape}}
                            ]
                        \addplot table [x=imbfactor, y=Tent]{\resultsclassrebalancingresnetbnfour};
                        \addplot table [x=imbfactor,y=DOT]{\resultsclassrebalancingresnetbnfour};
                        \addplot table [x=imbfactor, y=SAR]{\resultsclassrebalancingresnetbnfour};
                            \legend{Tent, DOT, SAR}
            
                        \end{axis}
                    \end{tikzpicture}        
                }
                \caption{ResNet50-BN-4}
            \end{subfigure}%
            \hfill
            \begin{subfigure}{0.33\columnwidth}
                \resizebox{\columnwidth}{!}{
                    \begin{tikzpicture}
                        \begin{axis}[
                            xlabel=Imbalance Factor,
                            ylabel=Accuracy (\%),
                            xmin=1, xmax=6000,
                            xtick=data,
                            xtick={1, 1000 ,2000, 3000, 4000 ,5000, 6000},
                            xticklabels={1, 1K ,2K, 3K, 4K ,5K, $\infty$},
                            ymin=10, ymax=50,
                            ytick={10,20,30,40,50,60,70,80,90,100},
                            legend columns=3,                         
                            legend style={
                                        draw=none,
                                        /tikz/every even column/.style={
                                         column sep=5pt,
                                        },
                                        at={(axis cs:6000,56)},anchor=north east,nodes={scale=1.7, transform shape}}
                                ]
                        \addplot table [x=imbfactor, y=Tent]{\resultsclassrebalancingresnetgnfour};
                        \addplot table [x=imbfactor,y=DOT]{\resultsclassrebalancingresnetgnfour};
                        \addplot table [x=imbfactor, y=SAR]{\resultsclassrebalancingresnetgnfour};
                            \legend{Tent, DOT, SAR}
            
                        \end{axis}
                    \end{tikzpicture}
                }
                    \caption{ResNet50-GN-4}
                \end{subfigure}%
                \hfill
                \begin{subfigure}{0.33\columnwidth}         
                    \resizebox{\columnwidth}{!}{
                        \begin{tikzpicture}
                            \begin{axis}[
                                xlabel=Imbalance Factor,
                                ylabel=Accuracy (\%),
                                xmin=1, xmax=6000,
                                xtick=data,
                                xtick={1, 1000 ,2000, 3000, 4000 ,5000, 6000},
                                xticklabels={1, 1K ,2K, 3K, 4K ,5K, $\infty$},
                                ymin=40, ymax=65,
                                ytick={10,20,30,40,50,60,70,80,90,100},
                                legend columns=3,                         
                                legend style={
                                        draw=none,
                                        /tikz/every even column/.style={
                                         column sep=5pt,
                                            },
                                            at={(axis cs:6000,69)},anchor=north east,nodes={scale=1.7, transform shape}}
                                ]
                                \addplot table [x=imbfactor, y=Tent]{\resultsclassrebalancingvitbasefour};
                                \addplot table [x=imbfactor,y= DOT]{\resultsclassrebalancingvitbasefour};
                                \addplot table [x=imbfactor, y=SAR]{\resultsclassrebalancingvitbasefour};
                                \legend{Tent, DOT, SAR}
                
                            \end{axis}
                        \end{tikzpicture}
                    }
                    \caption{VitBase-LN-4}
                \end{subfigure}
            \medskip
            \begin{subfigure}{0.33\columnwidth}
                \centering            
                \resizebox{\columnwidth}{!}{
                    \begin{tikzpicture}
                        \begin{axis}[
                            xlabel=Imbalance Factor,
                            ylabel=Accuracy (\%),
                            xmin=1, xmax=6000,
                            xtick=data,
                            xtick={1, 1000 ,2000, 3000, 4000 ,5000, 6000},
                            xticklabels={1, 1K ,2K, 3K, 4K ,5K, $\infty$},
                            ymin=0, ymax=60,
                            ytick={10,20,30,40,50,60,70,80,90,100},
                            legend columns=3,                         
                            legend style={
                                        draw=none,
                                        /tikz/every even column/.style={
                                         column sep=5pt,
                                        },
                                        at={(axis cs:6000,69)},anchor=north east,nodes={scale=1.7, transform shape}}
                            ]
                        \addplot table [x=imbfactor, y=Tent]{\resultsclassrebalancingresnetbntwo};
                        \addplot table [x=imbfactor,y=DOT]{\resultsclassrebalancingresnetbntwo};
                        \addplot table [x=imbfactor, y=SAR]{\resultsclassrebalancingresnetbntwo};
                            \legend{Tent, DOT, SAR}
            
                        \end{axis}
                    \end{tikzpicture}        
                }
                \caption{ResNet50-BN-2}
            \end{subfigure}%
            \hfill
            \begin{subfigure}{0.33\columnwidth}
                \resizebox{\columnwidth}{!}{
                    \begin{tikzpicture}
                        \begin{axis}[
                            xlabel=Imbalance Factor,
                            ylabel=Accuracy (\%),
                            xmin=1, xmax=6000,
                            xtick=data,
                            xtick={1, 1000 ,2000, 3000, 4000 ,5000, 6000},
                            xticklabels={1, 1K ,2K, 3K, 4K ,5K, $\infty$},
                            ymin=10, ymax=50,
                            ytick={10,20,30,40,50,60,70,80,90,100},
                        legend columns=3,                         
                        legend style={
                                    draw=none,
                                    /tikz/every even column/.style={
                                     column sep=5pt,
                                    },
                                    at={(axis cs:6000,56)},anchor=north east,nodes={scale=1.7, transform shape}}
                            ]
                        \addplot table [x=imbfactor, y=Tent]{\resultsclassrebalancingresnetgntwo};
                        \addplot table [x=imbfactor,y=DOT]{\resultsclassrebalancingresnetgntwo};
                        \addplot table [x=imbfactor, y=SAR]{\resultsclassrebalancingresnetgntwo};
                            \legend{Tent, DOT, SAR}
            
                        \end{axis}
                    \end{tikzpicture}
                }
                    \caption{ResNet50-GN-2}
                \end{subfigure}%
                \hfill
                \begin{subfigure}{0.33\columnwidth}         
                    \resizebox{\columnwidth}{!}{
                        \begin{tikzpicture}
                            \begin{axis}[
                                xlabel=Imbalance Factor,
                                ylabel=Accuracy (\%),
                                xmin=1, xmax=6000,
                                xtick=data,
                                xtick={1, 1000 ,2000, 3000, 4000 ,5000, 6000},
                                xticklabels={1, 1K ,2K, 3K, 4K ,5K, $\infty$},
                                ymin=40, ymax=65,
                                ytick={10,20,30,40,50,60,70,80,90,100},
                                legend columns=3,                         
                                legend style={
                                        draw=none,
                                        /tikz/every even column/.style={
                                         column sep=5pt,
                                            },
                                            at={(axis cs:6000,69)},anchor=north east,nodes={scale=1.7, transform shape}}
                                ]
                                \addplot table [x=imbfactor, y=Tent]{\resultsclassrebalancingvitbasetwo};
                                \addplot table [x=imbfactor,y= DOT]{\resultsclassrebalancingvitbasetwo};
                                \addplot table [x=imbfactor, y=SAR]{\resultsclassrebalancingvitbasetwo};
                                \legend{Tent, DOT, SAR}
                
                            \end{axis}
                        \end{tikzpicture}
                    }
                    \caption{VitBase-LN-2}
                \end{subfigure}
            \medskip
            \begin{subfigure}{0.33\columnwidth}
                \centering            
                \resizebox{\columnwidth}{!}{
                    \begin{tikzpicture}
                        \begin{axis}[
                            xlabel=Imbalance Factor,
                            ylabel=Accuracy (\%),
                            xmin=1, xmax=6000,
                            xtick=data,
                            xtick={1, 1000 ,2000, 3000, 4000 ,5000, 6000},
                            xticklabels={1, 1K ,2K, 3K, 4K ,5K, $\infty$},
                            ymin=0, ymax=60,
                            ytick={10,20,30,40,50,60,70,80,90,100},
                            legend columns=3,                         
                            legend style={
                                        draw=none,
                                        /tikz/every even column/.style={
                                         column sep=5pt,
                                        },
                                        at={(axis cs:6000,69)},anchor=north east,nodes={scale=1.7, transform shape}}
                            ]
                        \addplot table [x=imbfactor, y=Tent]{\resultsclassrebalancingresnetbnone};
                        \addplot table [x=imbfactor,y=DOT]{\resultsclassrebalancingresnetbnone};
                        \addplot table [x=imbfactor, y=SAR]{\resultsclassrebalancingresnetbnone};
                            \legend{Tent, DOT, SAR}
            
                        \end{axis}
                    \end{tikzpicture}        
                }
                \caption{ResNet50-BN-1}
            \end{subfigure}%
            \hfill
            \begin{subfigure}{0.33\columnwidth}
                \resizebox{\columnwidth}{!}{
                    \begin{tikzpicture}
                        \begin{axis}[
                            xlabel=Imbalance Factor,
                            ylabel=Accuracy (\%),
                            xmin=1, xmax=6000,
                            xtick=data,
                            xtick={1, 1000 ,2000, 3000, 4000 ,5000, 6000},
                            xticklabels={1, 1K ,2K, 3K, 4K ,5K, $\infty$},
                            ymin=10, ymax=50,
                            ytick={10,20,30,40,50,60,70,80,90,100},
                        legend columns=3,                         
                        legend style={
                                    draw=none,
                                    /tikz/every even column/.style={
                                     column sep=5pt,
                                    },
                                    at={(axis cs:6000,56)},anchor=north east,nodes={scale=1.7, transform shape}}
                            ]
                        \addplot table [x=imbfactor, y=Tent]{\resultsclassrebalancingresnetgnone};
                        \addplot table [x=imbfactor,y=DOT]{\resultsclassrebalancingresnetgnone};
                        \addplot table [x=imbfactor, y=SAR]{\resultsclassrebalancingresnetgnone};
                            \legend{Tent, DOT, SAR}
            
                        \end{axis}
                    \end{tikzpicture}
                }
                    \caption{ResNet50-GN-1}
                \end{subfigure}%
                \hfill
                \begin{subfigure}{0.33\columnwidth}         
                    \resizebox{\columnwidth}{!}{
                        \begin{tikzpicture}
                            \begin{axis}[
                                xlabel=Imbalance Factor,
                                ylabel=Accuracy (\%),
                                xmin=1, xmax=6000,
                                xtick=data,
                                xtick={1, 1000 ,2000, 3000, 4000 ,5000, 6000},
                                xticklabels={1, 1K ,2K, 3K, 4K ,5K, $\infty$},
                                ymin=40, ymax=65,
                                ytick={10,20,30,40,50,60,70,80,90,100},
                            legend columns=3,                         
                            legend style={
                                        draw=none,
                                        /tikz/every even column/.style={
                                         column sep=5pt,
                                        },
                                        at={(axis cs:6000,69)},anchor=north east,nodes={scale=1.7, transform shape}}
                                ]
                                \addplot table [x=imbfactor, y=Tent]{\resultsclassrebalancingvitbaseone};
                                \addplot table [x=imbfactor,y= DOT]{\resultsclassrebalancingvitbaseone};
                                \addplot table [x=imbfactor, y=SAR]{\resultsclassrebalancingvitbaseone};
                                \legend{Tent, DOT, SAR}
                
                            \end{axis}
                        \end{tikzpicture}
                    }
                    \caption{VitBase-LN-1}
                \end{subfigure}
        \caption{\textbf{Impact of Imbalance Factor, Architecture, and Batch Size on classification accuracy of different methods on ImageNet-C.} On ResNet50-BN, the performance of all models decreases when the imbalance factor increases. On ResNet50-GN, DOT, and SAR are more efficient than Tent, but SAR is more stable with very small batch sizes and stronger imbalance factors. On VitBase-LN, Tent performs lower than DOT and SAR with a batch size $\>$ 4 and a moderate imbalance factor. However, DOT and SAR performance is dropping significantly for small batch sizes and strong imbalance factors. The number after the architecture in the legend is the batch size.}
        \label{fig:imbfactor}            
    \end{figure}
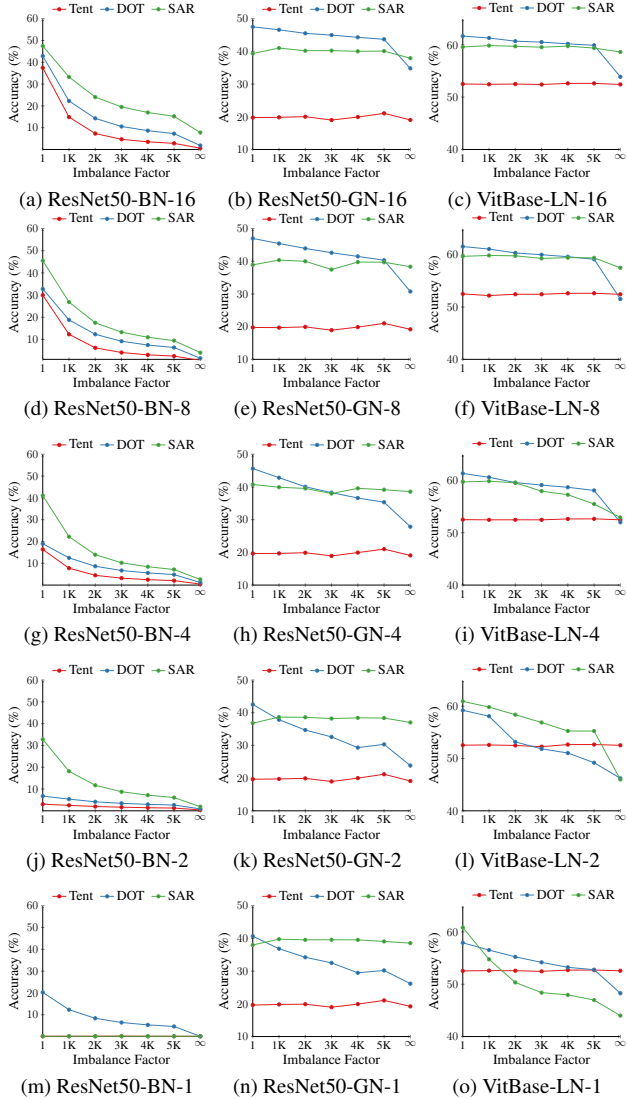

    To evaluate methods in regard to this problem, we consider two approaches.
    In the first one, we follow the setup proposed in \cite{niu2023towards}. In this setup, the online imbalanced label distribution shift is simulated by controlling the order of the input samples using a dataset generated using the following sampling strategy: a probability vector $Q_t(y) = [q_1, q_2, ..., q_K]$ is defined, where $t$ is a time step and $T$ is the total number of steps and is equal to $K$ the total number of classes, and $q_k=q_{max}$ if $k=t$ and $q_k=q_{min}\triangleq(1-q_{max})/(K-1)$ if $k\ne~t$. The ratio $q_{max}/q_{min}$ represents the imbalance ratio. For ImageNet-C, at each time step $t \in {1, 2, ..., T=K}$, 100 images are sampled using $Q_t(y)$ and so in total, the dataset contains 100x1000 images. An imbalance factor of 500000 is represented in Fig.~\ref{fig:imbfactor} as $\infty$ and represents a setup very close to the adaptation of the model one class after the other.
    Then, in a second approach, we investigate the evolution of the classification accuracy of different models simply in function of the batch size. We consider small batch sizes already as a factor of online class imbalance as not all classes can be present in the same batch.

    We compare three methods: i) Tent without any class rebalancing method is used as baseline. ii) SAR \cite{niu2023towards} is not a class rebalancing method per se but the sample selection method introduced in this work is presented as a way to address the class imbalance problem by the authors. iii) DOT is an adaptation of the class-wise reweighting method proposed in \cite{Cui2019ClassBalancedLB} adapted to the context of test-time adaptation in \cite{Zhao2023DELTADF}. The idea of DOT is to estimate the class frequencies in the test set by maintaining a momentum-based class-frequency vector $z \in \mathbb{R}^K$ where $K$ is the total number of classes, based on the prediction of the model of each sample seen previously. At inference time, each new sample receives a weight in function of its pseudo label and the current $z$ vector. A sample belonging to a rare class will receive a higher weight than a sample from a class seen more often. The DOT algorithm is detailed in the supplementary material.

    \paragraph{Experimental results} In Fig.~\ref{fig:imbfactor}, we can observe the following: i) On the ResNet50-BN architecture, the performance of all methods and for all batch sizes is dropping when the imbalance factor is increasing. Batch normalization does not seem to be a suitable normalization method when the test set is unbalanced ii) The performances of Tent and SAR are more stable when the imbalance factor varies on the ResNet50-GN architecture. On this architecture, DOT is the most performing method when the batch size is still high and the imbalance factor is still low. However, DOT performance is dropping drastically when the batch size becomes very small or the imbalance factor is very high. iii) Best performances are obtained by the VitBase-LN architecture. Performances are stable for all methods when the imbalance factor increases for a batch size of 16 or 8 but decrease when the imbalance factor increases for lower batch sizes.
    Our main takeaways from Fig.~\ref{fig:imbfactor} are that group normalization and layer normalization are less sensitive than batch normalization to imbalance classes and that even if DOT and SAR are both performing better than Tent, the sample selection of SAR yields more stable performances in the case of small batch sizes and stronger class imbalance factor.

\begin{figure}[!tp]
        \centering
        \begin{subfigure}{0.33\columnwidth}
            \centering            
            \resizebox{\columnwidth}{!}{
                \begin{tikzpicture}
                    \begin{axis}[
                        xlabel=Batch Size,
                        ylabel=Accuracy (\%),
                        xmin=1, xmax=16,
                        xtick=data,
                        x dir=reverse,
                        ymin=0, ymax=60,
                        ytick={10,20,30,40,50,60,70,80,90,100},
                        legend columns=3,                         
                        legend style={
                                    draw=none,
                                    /tikz/every even column/.style={
                                     column sep=5pt,
                                    },
                                    at={(axis cs:2,69)},anchor=north east,nodes={scale=1.7, transform shape}}
                        ]
                        \addplot table [x=bs, y=Tent]{\resultsclassrebalancinbsgresnetbn};
                        \addplot table [x=bs, y=DOT]{\resultsclassrebalancinbsgresnetbn};
                        \addplot table [x=bs, y=SAR]{\resultsclassrebalancinbsgresnetbn};
                        \legend{Tent, DOT, SAR}
        
                    \end{axis}
                \end{tikzpicture}        
            }
            \caption{ResNet50-BN}
        \end{subfigure}%
        \hfill
        \begin{subfigure}{0.33\columnwidth}
            \resizebox{\columnwidth}{!}{
                \begin{tikzpicture}
                    \begin{axis}[
                        xlabel=Batch Size,
                        ylabel=Accuracy (\%),
                        xmin=1, xmax=16,
                        xtick=data,
                        x dir=reverse,
                        ymin=10, ymax=50,
                        ytick={10,20,30,40,50,60,70,80,90,100},
                        legend columns=3,                         
                        legend style={
                                    draw=none,                            
                                    /tikz/every even column/.style={
                                     column sep=5pt,
                                    },
                                    at={(axis cs:2,56)},anchor=north east,nodes={scale=1.7, transform shape}}
                        ]
                        \addplot table [x=bs, y=Tent]{\resultsclassrebalancingbsresnetgn};
                        \addplot table [x=bs, y=DOT]{\resultsclassrebalancingbsresnetgn};
                        \addplot table [x=bs, y=SAR]{\resultsclassrebalancingbsresnetgn};
                        \legend{Tent, DOT, SAR}
        
                    \end{axis}
                \end{tikzpicture}
            }
            \caption{ResNet50-GN}
        \end{subfigure}%
        \hfill
        \begin{subfigure}{0.33\columnwidth}         
            \resizebox{\columnwidth}{!}{
                \begin{tikzpicture}
                    \begin{axis}[
                        xlabel=Batch Size,
                        ylabel=Accuracy (\%),
                        xmin=1, xmax=16,
                        xtick=data,
                        x dir=reverse,
                        ymin=40, ymax=65,
                        ytick={10,20,30,40,50,60,70,80,90,100},
                        legend columns=3,                         
                        legend style={
                                    draw=none,                            
                                    /tikz/every even column/.style={
                                     column sep=5pt,
                                    },
                                    at={(axis cs:2,69)},anchor=north east,nodes={scale=1.7, transform shape}}
                        ]
                        \addplot table [x=bs, y=Tent]{\resultsclassrebalancingbsvitbaseln};
                        \addplot table [x=bs, y=DOT]{\resultsclassrebalancingbsvitbaseln};
                        \addplot table [x=bs, y=SAR]{\resultsclassrebalancingbsvitbaseln};
                        \legend{Tent, DOT, SAR}
        
                    \end{axis}
                \end{tikzpicture}
            }
            \caption{VitBase-LN}
        \end{subfigure}
        \caption{\textbf{Impact of Architecture and Batch Size on the classification accuracy of different methods on ImageNet-C.} Batch-agnostic normalizations like group or layer normalization are more suitable to handle small batch sizes. Moreover, in this scenario, the class rebalancing method DOT is performing better than the sample selection method of SAR.}
        \label{fig:imbfactorbs}               
    \end{figure}
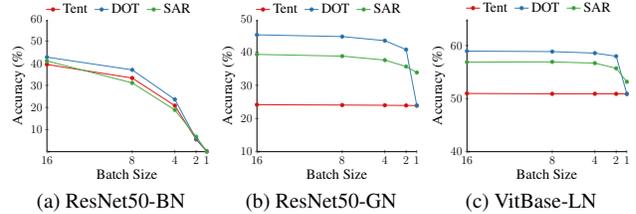

    In Fig.~\ref{fig:imbfactorbs}, we observe that the performance of all methods on ResNet50-BN is dropping when the batch size decreases. On ResNet50-GN and VitBase-LN, the classification accuracy remains stable when the batch size decreases for all models, DOT yielding the best results except when the batch size is 1. This particular case is explained in the next paragraph.
    Our main takeaways from Fig.~\ref{fig:imbfactorbs} are that architectures with group or layer normalization are more suitable to handle small batch sizes and that the class rebalancing method DOT is performing better than the sample selection method SAR for small batch sizes greater than 1.
    
    \pgfplotstableread[row sep=\\,col sep=&]{
        imbfactor & Tent & DOT & SAR\\
        1 & 37.44 & 42.88 & 47.48\\
        1000 & 14.87 & 22.26 & 33.26\\
        2000 & 7.30 & 14.27 & 24.03\\
        3000 & 4.68 & 10.54 & 19.51\\
        4000 & 3.50 & 8.65 & 16.98\\
        5000 & 2.80 & 7.26 & 15.20\\
        6000 & 0.61 & 1.82 & 7.78\\
    }\resultsclassrebalancingresnetbnsixteen

    \pgfplotstableread[row sep=\\,col sep=&]{
        imbfactor & Tent & SAR & DOT \\
        1 & 19.75 & 39.33 & 47.50	\\
        1000 & 19.83 & 41.01 & 46.60	\\
        2000 & 20.03 & 40.20 & 45.51	\\
        3000 & 19.00 & 40.24 & 44.98	\\
        4000 & 19.92 & 40.02 & 44.30	\\
        5000 & 21.05 & 40.08 & 43.69	\\
        6000 & 19.03 & 37.91 & 34.80	\\
    }\resultsclassrebalancingresnetgnsixteen

    \pgfplotstableread[row sep=\\,col sep=&]{
        imbfactor & Tent & SAR & DOT \\
        1 & 52.52 & 59.59 & 61.67 \\
        1000 & 52.45 & 59.86 & 61.31 \\
        2000 & 52.51 & 59.72 & 60.70 \\
        3000 & 52.42 & 59.55 & 60.52 \\
        4000 & 52.63 & 59.75 & 60.19 \\
        5000 & 52.63 & 59.38 & 59.91 \\
        6000 & 52.43 & 58.64 & 53.89 \\
    }\resultsclassrebalancingvitbasesixteen

    \pgfplotstableread[row sep=\\,col sep=&]{
        imbfactor & Tent & DOT & SAR \\
        1 & 29.96 & 32.77 & 45.50 \\
        1000 & 12.29 & 18.80 & 26.85 \\
        2000 & 6.20 & 12.30 & 17.50 \\
        3000 & 4.08 & 9.17 & 13.25 \\
        4000 & 3.03 & 7.46 & 10.96 \\
        5000 & 2.47 & 6.35 & 9.46 \\
        6000 & 0.52 & 1.49 & 4.02 \\
    }\resultsclassrebalancingresnetbneight

    \pgfplotstableread[row sep=\\,col sep=&]{
        imbfactor & Tent & SAR & DOT \\
        1 & 19.75 & 38.90 & 46.97 \\
        1000 & 19.71 & 40.36 & 45.42 \\
        2000 & 19.93 & 39.98 & 43.93 \\
        3000 & 18.92 & 37.47 & 42.60 \\
        4000 & 19.86 & 39.76 & 41.48 \\
        5000 & 21.01 & 39.70 & 40.30 \\
        6000 & 19.18 & 38.34 & 30.75 \\
    }\resultsclassrebalancingresnetgneight

    \pgfplotstableread[row sep=\\,col sep=&]{
        imbfactor & Tent & SAR & DOT \\
        1 & 52.51 & 59.69 & 61.56 \\
        1000 & 52.19 & 59.86 & 61.08 \\
        2000 & 52.44 & 59.79 & 60.34 \\
        3000 & 52.44 & 59.29 & 60.00 \\
        4000 & 52.62 & 59.44 & 59.62 \\
        5000 & 52.63 & 59.40 & 59.13 \\
        6000 & 52.43 & 57.50 & 51.53 \\
    }\resultsclassrebalancingvitbaseeight

    \pgfplotstableread[row sep=\\,col sep=&]{
        imbfactor & Tent & DOT & SAR\\
        1 & 16.35 & 18.97 & 41.08 \\
        1000 & 7.75 & 12.49 & 22.20 \\
        2000 & 4.48 & 8.64 & 13.95 \\
        3000 & 3.18 & 6.68 & 10.23 \\
        4000 & 2.48 & 5.59 & 8.42 \\
        5000 & 2.05 & 4.80 & 7.15 \\
        6000 & 0.46 & 1.19 & 2.56 \\
    }\resultsclassrebalancingresnetbnfour

    \pgfplotstableread[row sep=\\,col sep=&]{
        imbfactor & Tent & SAR & DOT \\
        1 & 19.64 & 40.76 & 45.64 \\
        1000 & 19.69 & 39.93 & 42.84 \\
        2000 & 19.90 & 39.55 & 40.07 \\
        3000 & 18.92 & 37.99 & 38.26 \\
        4000 & 19.94 & 39.58 & 36.63 \\
        5000 & 20.99 & 39.16 & 35.33 \\
        6000 & 19.06 & 38.57 & 27.85 \\
    }\resultsclassrebalancingresnetgnfour

    \pgfplotstableread[row sep=\\,col sep=&]{
        imbfactor & Tent & SAR & DOT \\
        1 & 52.51 & 59.71 & 61.34 \\
        1000 & 52.47 & 59.86 & 60.60 \\
        2000 & 52.48 & 59.55 & 59.59 \\
        3000 & 52.47 & 57.95 & 59.11 \\
        4000 & 52.64 & 57.27 & 58.69 \\
        5000 & 52.66 & 55.49 & 58.09 \\
        6000 & 52.48 & 52.92 & 51.99 \\
    }\resultsclassrebalancingvitbasefour

    \pgfplotstableread[row sep=\\,col sep=&]{
        imbfactor & Tent & DOT & SAR \\
        1 & 03.03 & 6.75 & 32.78 \\
        1000 & 2.51 & 5.36 & 18.15 \\
        2000 & 1.97 & 4.14 & 11.70 \\
        3000 & 1.66 & 3.44 & 8.74 \\
        4000 & 1.41 & 2.95 & 7.18 \\
        5000 & 1.25 & 2.66 & 06.08 \\
        6000 & 0.41 & 0.88 & 1.89 \\
    }\resultsclassrebalancingresnetbntwo

    \pgfplotstableread[row sep=\\,col sep=&]{
        imbfactor & Tent & SAR & DOT \\
        1 & 19.65 & 36.80 & 42.52 \\
        1000 & 19.75 & 38.66 & 37.85 \\
        2000 & 19.92 & 38.59 & 34.70 \\
        3000 & 18.97 & 38.21 & 32.60 \\
        4000 & 19.99 & 38.43 & 29.33 \\
        5000 & 21.18 & 38.37 & 30.32 \\
        6000 & 19.11 & 37.02 & 23.83 \\
    }\resultsclassrebalancingresnetgntwo

    \pgfplotstableread[row sep=\\,col sep=&]{
        imbfactor & Tent & DOT & SAR\\
        1 & 52.54 & 59.21 & 60.93 \\
        1000 & 52.59 & 58.08 & 59.82 \\
        2000 & 52.49 & 53.17 & 58.37 \\
        3000 & 52.26 & 51.84 & 56.88 \\
        4000 & 52.66 & 51.05 & 55.26 \\
        5000 & 52.67 & 49.19 & 55.24 \\
        6000 & 52.52 & 46.21 & 45.99 \\
    }\resultsclassrebalancingvitbasetwo

    \pgfplotstableread[row sep=\\,col sep=&]{
        imbfactor & Tent & SAR & DOT \\
        1 & 0.14 & 0.13 & 20.25 \\
        1000 & 0.12 & 0.14 & 12.32 \\
        2000 & 0.14 & 0.13 & 8.37 \\
        3000 & 0.18 & 0.15 & 6.43 \\
        4000 & 0.13 & 0.14 & 5.35 \\
        5000 & 0.12 & 0.14 & 4.62 \\
        6000 & 0.11 & 0.14 & 0.11 \\
    }\resultsclassrebalancingresnetbnone

    \pgfplotstableread[row sep=\\,col sep=&]{
        imbfactor & Tent & SAR & DOT \\
        1 & 19.63 & 38.00 & 40.68 \\
        1000 & 19.81 & 39.78 & 36.85 \\
        2000 & 19.89 & 39.57 & 34.24 \\
        3000 & 18.97 & 39.60 & 32.52 \\
        4000 & 19.92 & 39.56 & 29.48 \\
        5000 & 21.05 & 39.08 & 30.22 \\
        6000 & 19.22 & 38.57 & 26.17 \\
    }\resultsclassrebalancingresnetgnone

    \pgfplotstableread[row sep=\\,col sep=&]{
        imbfactor & Tent & SAR & DOT \\
        1 & 52.54 & 60.84 & 57.92   \\
        1000 & 52.60 & 54.76 & 56.53 \\
        2000 & 52.58 & 50.34 & 55.23 \\
        3000 & 52.46 & 48.39 & 54.18 \\
        4000 & 52.70 & 47.97 & 53.23 \\
        5000 & 52.70 & 46.97 & 52.78 \\
        6000 & 52.57 & 44.00 & 48.27 \\
    }\resultsclassrebalancingvitbaseone

    \pgfplotstableread[row sep=\\,col sep=&]{
        bs & Tent & SAR & DOT \\
        16 & 39.42595461 & 41.02875498 & 42.74542101 \\
        8 & 33.30039919 & 31.09653257 & 36.99088813 \\
        4 & 20.81062171 & 18.89511064 & 23.6259105 \\
        2 & 5.530888753 & 6.777555386 & 5.790755386 \\
        1 & 0.1446222183 & 0.1430666611 & 0.1448444406 \\
    }\resultsclassrebalancinbsgresnetbn

    \pgfplotstableread[row sep=\\,col sep=&]{
        bs & Tent & SAR & DOT \\
        16 & 24.15248823 & 39.32324342 & 45.22039905 \\
        8 & 24.05693284 & 38.80377697 & 44.69835413 \\
        4 & 23.98795491 & 37.61137733 & 43.47004314 \\
        2 & 23.92244371 & 35.65942122 & 40.77395439 \\
        1 & 23.90155499 & 33.85733259 & 23.90604384 \\
    }\resultsclassrebalancingbsresnetgn

    \pgfplotstableread[row sep=\\,col sep=&]{
        bs & Tent & SAR & DOT \\
        16 & 50.96533203 & 56.8703988 & 58.94919857 \\
        8 & 50.8992431 & 56.91653205 & 58.85537643 \\
        4 & 50.90853221 & 56.68528756 & 58.57079926 \\
        2 & 50.8908435 & 55.71159854 & 57.98337648 \\
        1 & 50.88835445 & 53.15897615 & 50.89035405 \\
    }\resultsclassrebalancingbsvitbaseln
    
    \paragraph{Single point learning for DOT method}
        In Fig.~\ref{fig:imbfactorbs}, we observe that in the specific case of batch size 1, the performance of DOT drops to the level of Tent. This is because in DOT, the weight of each sample in a batch is normalized by the sum of all weights of this batch. So, when the batch size is 1, the sum of the weights of the batch is equal to the weight of the single sample of the batch. Thus, the normalization of the weight of this single sample by the sum of all weights of the batch gives a weight of 1 and brings back to the same loss formulation as Tent.
        To address this issue, we propose to approximate the weight of a single sample in this particular case as if it was part of a bigger batch of size N. This approach does not require any additional processing time as we can still infer the class of an input test sample immediately and it is very cheap in terms of memory as we do not need to save any sample in a queue but just the weights of the N previous samples, which are only scalars. 
        In Tab.~\ref{tab:cr_bs1}, we analyze the impact of a buffer of different sizes on Tent performance on different architecture when the batch size is 1. We can see that an additional buffer of size 2 yields a significant performance improvement. Higher buffers yield no additional improvement on ResNet50-BN and a performance decrease on ResNet50-Gn and VitBase-LN. We assume that they lead to sample weights that are too noisy.

        \begin{table}[!tp]
            \centering
            \setlength\tabcolsep{1pt}
            \renewcommand*{\arraystretch}{1.1}
            \scriptsize
            \begin{tabular}{l|c|c|c|c|c}
                BatchSize=1 &DOT &DOT+buff=2&DOT+buff=4&DOT+buff=8&DOT+buff=16\\
                \hline
                ResNet50-BN &0.14\tiny$\pm$0.00 &\textbf{20.31}\tiny$\pm$0.02 & 20.31\tiny$\pm$0,02 & 20.31\tiny$\pm$0.02 & 20.31\tiny$\pm$0.02 \\
                ResNet50-GN &23.91\tiny$\pm$0,60 &\textbf{38.94}\tiny$\pm$0.03 & 38.32\tiny$\pm$0,06 & 36.23\tiny$\pm$0.03 & 34.13\tiny$\pm$0.02 \\
                VitBase-LN &50.89\tiny$\pm$0.00 &\textbf{54.15}\tiny$\pm$0.03 & 50.56\tiny$\pm$0,04 & 46.39\tiny$\pm$0.01 & 42.13\tiny$\pm$0.06 \\
            \end{tabular}
            \caption{\textbf{Impact of Additional Buffer on Tent performance on different architecture on ImageNet-C in the single point learning scenario.} An additional buffer of size 2 yields a significant performance improvement. Higher buffer sizes can lead to noisy sample weights and yield no additional improvement on ResNet50-BN or a performance decrease on ResNet50-GN and VitBase-LN.}
            \label{tab:cr_bs1}
        \end{table}   

\section{Sample selection}
    \label{sec:sampleselection}
    In the previous sections, we explored standard mechanisms to address covariate shift (through normalization) and label shift (through class rebalancing). In this section, we go one step further and explore mechanisms that cast TTA as a noisy learning problem. In particular, we explore the sample selection method first proposed in \cite{Niu2022EfficientTM} and analyzed more thoroughly after in \cite{niu2023towards}. The main idea of this method is to select only reliable samples for the model adaptation. Indeed, in \cite{niu2023towards}, authors show that samples with high entropy are more likely to have a strong and noisy gradient potentially harmful to the model performance. Furthermore, low-entropy samples contribute more to the model adaptation than high-entropy ones. However, there is no easy way to directly filter out samples with a strong gradient from the optimization process. So, instead, an entropy-based filtering method was proposed. More precisely, a threshold entropy $E_0$ is defined as the maximum entropy $\log{K}$ multiplied by a factor $F$, which is a scalar with a value between 0 and 1, 1 meaning no selection at all. All samples with an entropy below this threshold $F\log{K}$ are kept whereas the others are discarded when computing the loss value to update the model.
    Formally, this filtering method can be expressed as a sample selection function $S$:
    \begin{equation}
        \label{eq:SAR}
        S(x) = \ROMAN{II}_{\{E(x;\Theta)<E_0\}}(x)
    \end{equation}
    where $\ROMAN{II}_{\{.\}}(.)$ is an indicator function, $E(x;\Theta)$ is the entropy of sample $x$, and $E_0$ is a threshold predefined as:
    \begin{equation}
        \label{eq:E0}
        E_0 = F \log{K}
    \end{equation}
    where $K$ is the total number of classes in the dataset and $F$ is a real number in $[0;1]$.

\begin{figure}[!tp]
            \centering
            \begin{subfigure}{0.33\columnwidth}
                \centering            
                \resizebox{\columnwidth}{!}{
                    \begin{tikzpicture}
                        \begin{axis}[
                            xlabel=Entropy Factor F,
                            ylabel=Accuracy (\%),
                            xmin=0, xmax=1,
                            xtick=data,
                            ymin=20, ymax=60,
                            legend columns=3,                         
                            legend style={
                                        draw=none,                            
                                        /tikz/every even column/.style={
                                         column sep=0pt,
                                        },
                                        at={(axis cs:1,65)},anchor=north east,nodes={scale=1, transform shape}}
                            ]
                            \addplot table [x=upb, y=ResNet50-BN]{\resultssampleselectbssixteen};
                            \addplot table [x=upb, y=ResNet50-GN]{\resultssampleselectbssixteen};
                            \addplot table [x=upb, y=VitBase]{\resultssampleselectbssixteen};
                            \node (mark) [draw, red, circle, minimum size = 5pt, inner sep=2pt, thick] at (axis cs: 0.4, 41.18) {};
                            \node (mark) [draw, red, circle, minimum size = 5pt, inner sep=2pt, thick] at (axis cs: 0.2, 33.45) {};
                            \node (mark) [draw, red, circle, minimum size = 5pt, inner sep=2pt, thick] at (axis cs: 0.3, 55.77) {};            
                        \end{axis}
                    \end{tikzpicture}        
                }
                \caption{Batch Size=16}
            \end{subfigure}%
            \hfill
            \begin{subfigure}{0.33\columnwidth}
                \centering            
                \resizebox{\columnwidth}{!}{
                    \begin{tikzpicture}
                        \begin{axis}[
                            xlabel=Entropy Factor F,
                            ylabel=Accuracy (\%),
                            xmin=0, xmax=1,
                            xtick=data,
                            ymin=20, ymax=60,
                            legend columns=3,                         
                            legend style={
                                        draw=none,                            
                                        /tikz/every even column/.style={
                                         column sep=0pt,
                                        },
                                        at={(axis cs:1,65)},anchor=north east,nodes={scale=1, transform shape}}
                            ]
                            \addplot table [x=upb, y=ResNet50-BN]{\resultssampleselectbseight};
                            \addplot table [x=upb, y=ResNet50-GN]{\resultssampleselectbseight};
                            \addplot table [x=upb, y=VitBase]{\resultssampleselectbseight};
                            \node (mark) [draw, red, circle, minimum size = 5pt, inner sep=2pt, thick] at (axis cs: 0.3, 35.64) {};
                            \node (mark) [draw, red, circle, minimum size = 5pt, inner sep=2pt, thick] at (axis cs: 0.2, 33.34) {};
                            \node (mark) [draw, red, circle, minimum size = 5pt, inner sep=2pt, thick] at (axis cs: 0.3, 55.79) {};            
                        \end{axis}
                    \end{tikzpicture}        
                }
                \caption{Batch Size=8}
            \end{subfigure}%
            \hfill
            \begin{subfigure}{0.33\columnwidth}
                \centering            
                \resizebox{\columnwidth}{!}{
                    \begin{tikzpicture}
                        \begin{axis}[
                            xlabel=Entropy Factor F,
                            ylabel=Accuracy (\%),
                            xmin=0, xmax=1,
                            xtick=data,
                            ymin=00, ymax=60,
                            legend columns=3,                         
                            legend style={
                                        draw=none,                            
                                        /tikz/every even column/.style={
                                         column sep=0pt,
                                        },
                                        at={(axis cs:1,68)},anchor=north east,nodes={scale=1, transform shape}}
                            ]
                            \addplot table [x=upb, y=ResNet50-BN]{\resultssampleselectbsfour};
                            \addplot table [x=upb, y=ResNet50-GN]{\resultssampleselectbsfour};
                            \addplot table [x=upb, y=VitBase]{\resultssampleselectbsfour};
                            \node (mark) [draw, red, circle, minimum size = 5pt, inner sep=2pt, thick] at (axis cs: 0.6, 22.65) {};
                            \node (mark) [draw, red, circle, minimum size = 5pt, inner sep=2pt, thick] at (axis cs: 0.2, 32.96) {};
                            \node (mark) [draw, red, circle, minimum size = 5pt, inner sep=2pt, thick] at (axis cs: 0.3, 55.16) {};            
                        \end{axis}
                    \end{tikzpicture}        
                }
                \caption{Batch Size=4}
            \end{subfigure}
            \medskip
            \begin{subfigure}{0.33\columnwidth}
                \centering            
                \resizebox{\columnwidth}{!}{
                    \begin{tikzpicture}
                        \begin{axis}[
                            xlabel=Entropy Factor F,
                            ylabel=Accuracy (\%),
                            xmin=0, xmax=1,
                            xtick=data,
                            ymin=0, ymax=60,
                            legend columns=3,                         
                            legend style={
                                        draw=none,                            
                                        /tikz/every even column/.style={
                                         column sep=0pt,
                                        },
                                        at={(axis cs:1,68)},anchor=north east,nodes={scale=1, transform shape}}
                            ]
                            \addplot table [x=upb, y=ResNet50-BN]{\resultssampleselectbstwo};
                            \addplot table [x=upb, y=ResNet50-GN]{\resultssampleselectbstwo};
                            \addplot table [x=upb, y=VitBase]{\resultssampleselectbstwo};
                            \node (mark) [draw, red, circle, minimum size = 5pt, inner sep=2pt, thick] at (axis cs: 0.7, 6.79) {};
                            \node (mark) [draw, red, circle, minimum size = 5pt, inner sep=2pt, thick] at (axis cs: 0.2, 31.98) {};
                            \node (mark) [draw, red, circle, minimum size = 5pt, inner sep=2pt, thick] at (axis cs: 0.3, 54.06) {};
                        \end{axis}
                    \end{tikzpicture}        
                }
                \caption{Batch Size=2}
            \end{subfigure}%
            \hspace*{0.005\textwidth}%
            \begin{subfigure}{0.33\columnwidth}
                \centering            
                \resizebox{\columnwidth}{!}{
                    \begin{tikzpicture}
                        \begin{axis}[
                            xlabel=Entropy Factor F,
                            ylabel=Accuracy (\%),
                            xmin=0, xmax=1,
                            xtick=data,
                            ymin=-2, ymax=60,
                            legend style={
                                        overlay,
                                        draw=none,                            
                                        /tikz/every even column/.style={
                                         column sep=0pt,
                                        },
                                        at={(1.03,0.5)},
                                        anchor=west,
                                        nodes={scale=1.8, transform shape}
                                        }
                            ]
                            \addplot table [x=upb, y=ResNet50-BN]{\resultssampleselectbsone};
                            \addplot table [x=upb, y=ResNet50-GN]{\resultssampleselectbsone};
                            \addplot table [x=upb, y=VitBase]{\resultssampleselectbsone};
                            \legend{ResNet50-BN, ResNet50-GN, VitBase-LN}
                            \node (mark) [draw, red, circle, minimum size = 5pt, inner sep=2pt, thick] at (axis cs: 1.0, 0.14) {};
                            \node (mark) [draw, red, circle, minimum size = 5pt, inner sep=2pt, thick] at (axis cs: 0.3, 32.36) {};
                            \node (mark) [draw, red, circle, minimum size = 5pt, inner sep=2pt, thick] at (axis cs: 0.4, 51.80) {};
                        \end{axis}
                    \end{tikzpicture}        
                }
                \caption{Batch Size=1}
            \end{subfigure}%
            \hspace*{0.05\textwidth}
            
            \caption{\textbf{Impact of Sample Selection and Architecture on classification accuracy of different methods on ImageNet-C}. The best results are circled in red. The optimal threshold varies in function of the architecture and the batch size and is lower for the smaller batch sizes than the values 0.5 or 0.4 for a batch size of 64 recommended in \cite{Niu2022EfficientTM}.}
            \label{fig:sampleselect}            
        \end{figure}
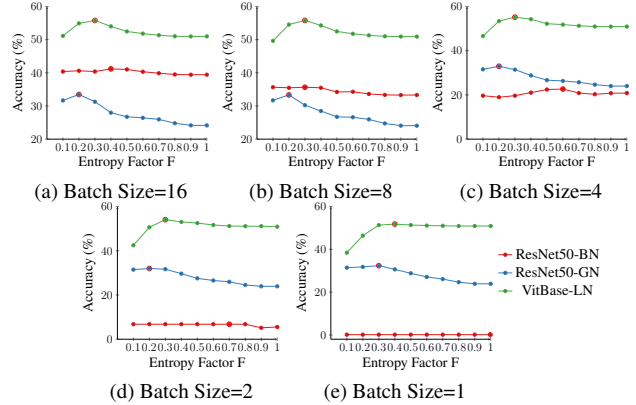

    \paragraph{Experimental results} In Fig.~\ref{fig:sampleselect}, we can see that fine-tuning the selection threshold via factor $F$ can lead to a significant increase in the performances in all cases. We also observe that in the case of smaller batch sizes, the optimal value for $F$ is smaller than the value of 0.5 recommended in \cite{niu2023towards} for a batch size of 64. Moreover, as mentioned in \cite{niu2023towards}, another advantage of this method is that it requires less computational power to perform the adaptation as fewer samples are used in the optimization. \eg for the Gaussian noise corruption, severity level 5, on ResNet50-GN and an entropy factor $F$ of 0.4, the model forward passes 50K samples but keep less than 13K after selection for the backward pass, which is only 26\% of the whole dataset. 

        \pgfplotstableread[row sep=\\,col sep=&]{
            upb & ResNet50-BN & ResNet50-GN & VitBase \\
            0.1 & 40.38 & 31.66 & 51.13\\
            0.2 & 40.61 & 33.45 & 54.89\\
            0.3 & 40.38 & 31.29 & 55.77\\
            0.4 & 41.18 & 27.98 & 53.99\\
            0.5 & 41.01 & 26.74 & 52.46\\
            0.6 & 40.33 & 26.43 & 51.82\\
            0.7 & 39.86 & 25.96 & 51.33\\
            0.8 & 39.50 & 24.80 & 51.00\\
            0.9 & 39.42 & 24.15 & 50.95\\
            1.0 & 39.43 & 24.15 & 50.97\\
        }\resultssampleselectbssixteen

        \pgfplotstableread[row sep=\\,col sep=&]{
            upb & ResNet50-BN & ResNet50-GN & VitBase \\
            0.1 & 35.64 & 31.69 & 49.63\\
            0.2 & 35.48 & 33.34 & 54.53\\
            0.3 & 35.64 & 30.23 & 55.79\\
            0.4 & 35.48 & 28.47 & 54.28\\
            0.5 & 34.22 & 26.73 & 52.47\\
            0.6 & 34.29 & 26.60 & 51.76\\
            0.7 & 33.62 & 25.96 & 51.28\\
            0.8 & 33.34 & 24.73 & 50.99\\
            0.9 & 33.28 & 24.07 & 50.92\\
            1.0 & 33.30 & 24.06 & 50.90\\
        }\resultssampleselectbseight

        \pgfplotstableread[row sep=\\,col sep=&]{
            upb & ResNet50-BN & ResNet50-GN & VitBase \\
            0.1 & 19.66 & 31.56 & 46.59\\
            0.2 & 18.94 & 32.96 & 53.35\\
            0.3 & 19.66 & 31.44 & 55.16\\
            0.4 & 21.04 & 28.80 & 54.23\\
            0.5 & 22.43 & 26.67 & 52.20\\
            0.6 & 22.65 & 26.33 & 51.79\\
            0.7 & 20.86 & 25.71 & 51.25\\
            0.8 & 20.28 & 24.66 & 50.92\\
            0.9 & 20.78 & 23.99 & 50.88\\
            1.0 & 20.81 & 23.99 & 50.91\\
        }\resultssampleselectbsfour

        \pgfplotstableread[row sep=\\,col sep=&]{
            upb & ResNet50-BN & ResNet50-GN & VitBase \\
            0.1 & 6.78 & 31.47 & 42.47\\
            0.2 & 6.78 & 31.98 & 50.61\\
            0.3 & 6.78 & 31.67 & 54.06\\
            0.4 & 6.78 & 29.65 & 53.05\\
            0.5 & 6.78 & 27.53 & 52.51\\
            0.6 & 6.78 & 26.56 & 51.66\\
            0.7 & 6.79 & 25.97 & 51.18\\
            0.8 & 6.79 & 24.55 & 51.10\\
            0.9 & 5.17 & 23.92 & 51.11\\
            1.0 & 5.53 & 23.92 & 50.89\\
        }\resultssampleselectbstwo

        \pgfplotstableread[row sep=\\,col sep=&]{
            upb & ResNet50-BN & ResNet50-GN & VitBase \\
            0.1 & 0.14 & 31.45 & 38.44\\
            0.2 & 0.14 & 31.80 & 46.39\\
            0.3 & 0.14 & 32.36 & 51.31\\
            0.4 & 0.14 & 30.64 & 51.80\\
            0.5 & 0.14 & 28.81 & 51.36\\
            0.6 & 0.14 & 27.12 & 51.11\\
            0.7 & 0.14 & 26.10 & 50.99\\
            0.8 & 0.14 & 24.65 & 50.90\\
            0.9 & 0.14 & 23.91 & 50.89\\
            1.0 & 0.14 & 23.90 & 50.89\\
        }\resultssampleselectbsone

\section{Calibration}
    \label{sec:calibration}
    In this section, we investigate the problem of network calibration in the context of Fully TTA. The calibration of classification networks is a measure of the confidence of the predictions. It is of utmost importance in the context of Fully TTA as it impacts directly the predictions entropy.
    Temperature scaling is one technique introduced in \cite{Guo2017OnCO} to improve the calibration of under- or overconfident neural networks by correcting the logits in the softmax function. 
    Formally, it is expressed as:
    \begin{equation}
        softmax_{\tau}(z)_i = \frac{e^{z_i/\tau}}{\sum_{j=1}^K e^{z_j/\tau}} 
    \end{equation}
    where $\tau$ is the temperature scaling factor, $z$ is the logits vector of an input sample, i is a class index and K is the total number of classes.
    A $\tau$ value above 1 will lead to a higher entropy with a flattened distribution of the model predictions whereas a $\tau$ value smaller than 1 will lead to a low entropy with a more peaky predictions distribution.
    In the context of test-time adaptation, \cite{Goyal2022TestTimeAV} shows that using temperature scaling improves the model accuracy after adaptation when using an entropy minimization-based method. \cite{lee2013pseudo} also shows that when meta-learning the optimal loss for test-time adaptation, the result is an entropy minimization loss with a temperature scaling factor.
    To determine the temperature scaling factor in our experiments, we follow \cite{Zhao2023DELTADF} in the way to select hyperparameters using the 4 Imagenet-C validation corruptions. For each network architecture, we select the temperature scaling factor $\tau$ for each validation corruption using a grid search on values between 0.5 and 1.5 with a step of 0.1 and keep the average of the 4 values.

    For the 3 network architectures considered, we obtain a temperature scaling factor of 1.2, which means that without correction, the models are too confident in their predictions.

    \paragraph{Experimental results} In Tab.~\ref{tab:temperature}, we observe that applying temperature scaling during adaptation leads to an increase in Tent performance on ResNet50-BN and VitBase-LN. On ResNet50-GN, the mean is slightly lower, but the standard deviation is significantly reduced, which means overall a better performance in terms of statistical significance.
    The performance increase is not very high when using temperature alone. However, we will see in Sec.~\ref{sec:trickcombi} that it leads to higher performance when combined with other tricks.

        \begin{table}[!t]
            \centering
            \setlength\tabcolsep{2pt}
            \renewcommand*{\arraystretch}{1.1}
            \scriptsize
            \begin{tabular}{l|c|c|c|c|c}
                &16 &8 &4 &2 &1 \\
                \hline
                ResNet50-BN &39.43\tiny$\pm$0.13 &33.30\tiny$\pm$0.04 &20.81\tiny$\pm$0.08 &5.53\tiny$\pm$0.01 &0.14\tiny$\pm$0.00 \\
                ResNet50-BN+ temp &\textbf{39,45}\tiny$\pm$0,06 &\textbf{33,86}\tiny$\pm$0,04 &\textbf{20,84}\tiny$\pm$0,07 &\textbf{6,11}\tiny$\pm$0,01 &\textbf{0,15}\tiny$\pm$0,00 \\
                \hline
                ResNet50-GN &\textbf{24,15}\tiny$\pm$0,55 &\textbf{24,00}\tiny$\pm$0,54 &\textbf{23,99}\tiny$\pm$0,56 &\textbf{23,92}\tiny$\pm$0,57 &\textbf{23,90}\tiny$\pm$0,58 \\
                ResNet50-GN+ temp &24,01\tiny$\pm$\textbf{0,17} &23,87\tiny$\pm$\textbf{0,17} &23,82\tiny$\pm$\textbf{0,15} &23,76\tiny$\pm$\textbf{0,19} &23,74\tiny$\pm$\textbf{0,19} \\
                \hline
                VitBase-LN &50,97\tiny$\pm$0,07 &50,90\tiny$\pm$0,04 &50,91\tiny$\pm$0,07 &50,89\tiny$\pm$0,06 &50,89\tiny$\pm$0.04 \\
                VitBase-LN + temp &\textbf{52,84}\tiny$\pm$0,27 &\textbf{52,81}\tiny$\pm$0,26 &\textbf{52,76}\tiny$\pm$0,26 &\textbf{52,76}\tiny$\pm$0,20 &\textbf{52,77}\tiny$\pm$0,22 \\
            \end{tabular}
            \caption{\textbf{Impact of Temperature on classification accuracy of Tent method performance on different architecture on ImageNet-C.} Using a temperature scaling factor increases the mean accuracy on ResNet50-BN and VitBase-LN. On ResNet50-GN, using temperature decreases slightly the mean classification accuracy but decreases also the standard deviation, which means that the model is better with respect to statistical significance.}
            \label{tab:temperature}            
        \end{table}    

        \pgfplotstableread[row sep=\\,col sep=&]{
            bs & ResNet50-BN & ResNet50-BN + temp & ResNet50-GN & ResNet50-GN + temp & VitBase-LN & VitBase-LN + temp\\
            16 & 39.42595461 & 39.44853261 & 24.15248823 & 24.00724396 & 50.96533203 & 52.84195438 \\
            8 & 33.30039919 & 32.85506615 & 24.05693284 & 23.87195484 & 50.8992431 & 52.81399846 \\
            4 & 20.81062171 & 20.23871076 & 23.98795491 & 23.8198216 & 50.90853221 & 52.7589759 \\
            2 & 5.530888753 & 6.111466527 & 23.92244371 & 23.75804382 & 50.8908435 & 52.763021 \\
            1 & 0.1446222183 & 0.1450222198 & 23.90155499 & 23.74431077 & 50.88835445 & 52.77039837 \\
        }\resultstemperature 

\section{Tricks combinations}
    \label{sec:trickcombi}

    In this section, we investigate the performance of Tent using different combinations of the tricks presented in the previous sections. For ResNet50-BN, we consider the usage of batch renormalization as an essential trick when dealing with very small batch sizes as presented in Sec.~\ref{sec:archnorm} and always integrate it in the different tricks combinations tested. In the ResNet50-BN section of Tab.~\ref{tab:tricks_combination}, we report first the results already presented in Fig.~\ref{fig:archinorm} to see the performance improvement with batch renormalization. Then we consider all the possible combinations of 2 of the tricks presented and finally, we consider the combination of all the tricks. For ResNet50-GN and VitBase-LN, we also present results considering all the possible combinations of 2 of the tricks presented previously and then combining all the tricks.

    \paragraph{Experimental results} In Tab.~\ref{tab:tricks_combination}, we observe that when using a ResNet50-BN network, the best pair of tricks is the class rebalancing method DOT combined with the entropy-based sample selection. The best results overall are obtained when using this pair with a temperature scaling factor, in other words when using all tricks together. In this case, compared to Tent, we obtain an average improvement of +17.08\% accuracy over all batch sizes.
    In the case of a ResNet50-GN architecture, the best pair of tricks is class rebalancing combined with the temperature scaling factor. Surprisingly, combining temperature scaling with sample selection is performing better than vanilla Tent but much lower than other pairs of tricks. We assume that as the temperature scaling is changing the entropy of the test samples, a finer tuning of the sample selection margin should be done to ensure that samples useful for the model adaptation are not discarded. The best performances are obtained using all tricks. In this case, we obtain an average improvement of +19.92\% accuracy over all batch sizes compared to Tent.
    When considering the VitBase-LN architecture, we can see that the two pairs of tricks class rebalancing and temperature and class rebalancing and sample selection are close over all the batch sizes and yield the best results of the pairs of tricks. The overall best results are obtained when combining all tricks. Doing this leads to an average improvement compared to Tent of +7.66\% over all batch sizes.
    Our main takeaway for this series of experiments is that the best results are obtained when combining all tricks (class rebalancing, sample selection, and temperature scaling), and this for the 3 architectures and the different batch sizes considered. Among the different architectures, VitBase-LN has the best classification accuracy when combining all the tricks and on all the batch sizes tested.

        \begin{table}[!tp]\centering
            \setlength\tabcolsep{2.5pt}
            \renewcommand*{\arraystretch}{1.1}
            \scriptsize
            \begin{tabular}{@{\hskip2pt}c@{\hskip2pt}|@{\hskip1pt}c@{\hskip1pt}@{\hskip1pt}c@{\hskip1pt}@{\hskip1pt}c@{\hskip1pt}@{\hskip1pt}c@{\hskip1pt}|c|c|c|c|c}
                 & \multicolumn{4}{c|}{Tent +} & \multicolumn{5}{c}{Batch Size}\\\cline{2-10}
                & BR & CR & SS & T & 16 & 8 & 4 & 2 & 1 \\
                \hline
                \multirow{6}{*}{\rot{ResNet50-BN}} &  & & & & 39.40\tiny$\pm$0.13 & 33.30\tiny$\pm$0.04 & 20.81\tiny$\pm$0.08 & 5.53\tiny$\pm$0.01 & 0.14\tiny$\pm$0.00\\
                & \checkmark &   &   &   & 43.26\tiny$\pm$0.01 & 41.39\tiny$\pm$0.06 & 37.72\tiny$\pm$0.05 & 30.84\tiny$\pm$0.04 & 20.25\tiny$\pm$0.01 \\
                & \checkmark & \checkmark &   & \checkmark & 45.89\tiny$\pm$0.06 & 43.70\tiny$\pm$0.05 & 39.17\tiny$\pm$0.05 & 31.44\tiny$\pm$0.04 & \textbf{20.31}\tiny$\pm$0.02 \\
                & \checkmark &  & \checkmark &  \checkmark & 45.17\tiny$\pm$0.26 & 43.03\tiny$\pm$0.11 & 39.02\tiny$\pm$0.07 & 31.60\tiny$\pm$0.05 & 20.26\tiny$\pm$0.01 \\
                & \checkmark &  \checkmark & \checkmark & & 46.57\tiny$\pm$0.07 & 44.46\tiny$\pm$0.01 & 39.95\tiny$\pm$0.01 & 31.65\tiny$\pm$0.01 & 20.30\tiny$\pm$0.02 \\
                & \checkmark & \checkmark & \checkmark & \checkmark & \textbf{46.90}\tiny$\pm$0.12 & \textbf{44.90}\tiny$\pm$0.09 & \textbf{40.42}\tiny$\pm$0.14 & \textbf{32.03}\tiny$\pm$0.05 & \textbf{20.31}\tiny$\pm$0.02 \\ 
                \hline
                \multirow{5}{*}{\rot{ResNet50-GN}} & & & & & 24.15\tiny$\pm$0.55 & 24.06\tiny$\pm$0.54 & 23.99\tiny$\pm$0.57 & 23.92\tiny$\pm$0.57 & 23.90\tiny$\pm$0.58\\
                & & \checkmark & & \checkmark & 46.35\tiny$\pm$0.07 & 45.89\tiny$\pm$0.09 & 44.77\tiny$\pm$0.01 & 42.07\tiny$\pm$0.03 & 39.31\tiny$\pm$0.64 \\
                & & & \checkmark & \checkmark & 26.85\tiny$\pm$0.17 & 27.34\tiny$\pm$0.55 & 29.03\tiny$\pm$0.59 & 30.19\tiny$\pm$0.20 & 27.20\tiny$\pm$0.48 \\
                & & \checkmark & \checkmark & & 45.78\tiny$\pm$0.09 & 45.31\tiny$\pm$0.11 & 44.21\tiny$\pm$0.01 & 41.33\tiny$\pm$0.01 & 38.94\tiny$\pm$0.03 \\
                & & \checkmark & \checkmark & \checkmark & \textbf{46.50}\tiny$\pm$0.05 & \textbf{46.07}\tiny$\pm$0.08 & \textbf{45.02}\tiny$\pm$0.01 & \textbf{42.32}\tiny$\pm$0.01 & \textbf{39.70}\tiny$\pm$0.04 \\
                \hline
                \multirow{5}{*}{\rot{VitBase-LN}} & & & & & 50.97\tiny$\pm$0.07 & 50.90\tiny$\pm$0.04 & 50.91\tiny$\pm$0.07 & 50.89\tiny$\pm$0.06 & 50.89\tiny$\pm$0.04\\
                & & \checkmark & & \checkmark & 59.26\tiny$\pm$0.03 & 59.20\tiny$\pm$0.02 & 58.97\tiny$\pm$0.04 & 58.52\tiny$\pm$0.05 & 54.68\tiny$\pm$0.03 \\
                & & & \checkmark & \checkmark & 57.59\tiny$\pm$0.44 & 58.11\tiny$\pm$0.14 & 57.88\tiny$\pm$0.09 & 57.02\tiny$\pm$0.10 & 55.10\tiny$\pm$0.07 \\
                & & \checkmark & \checkmark & & 59.31\tiny$\pm$0.06 & 59.22\tiny$\pm$0.04 & 58.96\tiny$\pm$0.00 & 57.51\tiny$\pm$0.78 & 54.15\tiny$\pm$0.03 \\
                & & \checkmark & \checkmark & \checkmark & \textbf{59.80}\tiny$\pm$0.07 & \textbf{59.77}\tiny$\pm$0.04 & \textbf{59.59}\tiny$\pm$0.03 & \textbf{59.04}\tiny$\pm$0.06 & \textbf{55.15}\tiny$\pm$0.03 \\
                
                \hline
                \multicolumn{10}{l}{\scriptsize{BR=BatchRenorm, T=Temperature, CR=Class Rebalancing, SS=Sample Selection}} \\
                
            \end{tabular}
            \caption{\textbf{Effect of Tricks Combination on model performance.} Best results are obtained when combining all tricks and this for the 3 architectures and the different batch sizes considered. Among the different architectures, VitBase-LN has the best classification accuracy in all the different setups.}
            \label{tab:tricks_combination}            
        \end{table}        

\section{Comparison to other methods and on other datasets}
    \label{sec:otherdatasets}
    
    In this final experimental section, we compare the performance of BoT (i.e. Tent with all the tricks presented in this article) to a vanilla Tent and 2 state-of-the-art methods, SAR \cite{niu2023towards} and Delta \cite{Zhao2023DELTADF}. This comparison is performed on different network architectures and different datasets: ResNet50-BN, ResNet50-GN, VitBase-LN for ImageNet-C, ImageNet-Rendition and ImageNet-Sketch, and ResNet101 for VisDA2017.
    
    \paragraph{Experimental results} In Tab.~\ref{tab:compmethods_imagenetc}, we can see that on the ImageNet-C dataset, BoT obtains better results than a vanilla Tent, and the two state-of-the-art methods for all the batch sizes considered. Interesting to see is the collapse of SAR performance for very small batch sizes (2 and 1) on ResNet50-BN that we do not observe with Delta due to the usage of batch renormalization. If the performance increase by using all the tricks is not significant on ResNet50-BN (+0.78\% accuracy on average versus Delta), it is much more noticeable on ResNet50-GN (+4.31\% accuracy on average versus Delta) and VitBase-LN (+1.53\% accuracy in average versus Delta).
    In Tab.~\ref{tab:compmethods_imagenetrendition}, we also observe that BoT performs the best in all cases. Interesting to note is that results are more stable over the different batch sizes with ResNet50-GN compared to ResNet50-BN, which is in line with observations from previous experiments. Delta performs better than SAR but worse than BoT. The performance increase of BoT compared to Delta is similar on ResNet50-BN and ResNet50-GN (respectively +0.85\% and +0.87\% accuracy) but reaches +1.23\% accuracy on VitBase-LN.
    In Tab.~\ref{tab:compmethods_imagenetsketch}, we make the same observations on ImageNet-Sketch as on the other ImageNet variants. ResNet50-BN performance drops when the batch size becomes small. In all cases, Delta performs better than SAR but not as good as BoT. BoT performs best in all cases. The performance increase of BoT versus Delta is +0.72\% accuracy on ResNet50-BN, +1.32\% accuracy on ResNet50-GN, and +1.03\% accuracy on VitBase-LN.
    In Tab.~\ref{tab:compmethods_visda2017}, we observe that also for the VisDA2017 dataset, results are in line with previous experiments. Delta performs better than Tent and SAR but not as well as BoT. The performance improvement of BoT versus Delta is +0.36\% accuracy on ResNet101.
        
    \begin{table}[!tp]
        \centering
        \setlength\tabcolsep{3pt}
        \scriptsize
        \begin{tabular}{c|c|c|c|c|c|c}
             & \multirow{2}{*}{Method} & \multicolumn{5}{c}{Batch Size}\\\cline{3-7}
             & & 16 & 8 & 4 & 2 & 1 \\
            \hline
            \multirow{4}{*}{\tiny\rot{ResNet50-BN}} & \>Tent & 39.43\tiny$\pm$0.13 & 33.30\tiny$\pm$0.04 & 20.81\tiny$\pm$0.08 & 5.53\tiny$\pm$0.01 & 0.14\tiny$\pm$0.00 \\
            & \>SAR & 41.02\tiny$\pm$0.29 & 31.10\tiny$\pm$0.08 & 18.90\tiny$\pm$0.04 & 6.78\tiny$\pm$0.00 & 0.14\tiny$\pm$0.00 \\
            & \>Delta & 46.33\tiny$\pm$0.78 & 43.67\tiny$\pm$0.05 & 39.16\tiny$\pm$0.04 & 31.26\tiny$\pm$0.05 & 20.25\tiny$\pm$0.01 \\
            & BoT & \textbf{46.90}\tiny$\pm$0.1 & \textbf{44.90}\tiny$\pm$0.09 & \textbf{40.42}\tiny$\pm$0.14 & \textbf{32.03}\tiny$\pm$0.05 & \textbf{20.31}\tiny$\pm$0.02 \\
            \hline
            \multirow{4}{*}{\tiny\rot{ResNet50-GN}} & Tent & 24.15\tiny$\pm$0.55 & 24.05\tiny$\pm$0.54 & 23.99\tiny$\pm$0.57 & 23.92\tiny$\pm$0.57 & 23.90\tiny$\pm$0.58 \\
            & SAR & 39.32\tiny$\pm$0.17 & 38.80\tiny$\pm$0.14 & 37.61\tiny$\pm$0.39 & 35.66\tiny$\pm$0.28 & 33.86\tiny$\pm$0.06 \\
            & Delta & 45.22\tiny$\pm$0.06 & 44.70\tiny$\pm$0.09 & 43.47\tiny$\pm$0.02 & 40.77\tiny$\pm$0.01 & 23.91\tiny$\pm$0.60 \\
            & BoT & \textbf{46.50}\tiny$\pm$0.05 & \textbf{46.07}\tiny$\pm$0.08 & \textbf{45.02}\tiny$\pm$0.01 & \textbf{42.32}\tiny$\pm$0.01 & \textbf{39.70}\tiny$\pm$0.04 \\
            \hline
            \multirow{4}{*}{\tiny\rot{VitBase-LN}} & Tent & 50.97\tiny$\pm$0.07 & 50.90\tiny$\pm$0.04 & 50.91\tiny$\pm$0.07 & 50.90\tiny$\pm$0.06 & 50.89\tiny$\pm$0.04 \\
            & SAR & 56.87\tiny$\pm$0.15 & 56.92\tiny$\pm$0.10 & 56.69\tiny$\pm$0.13 & 55.71\tiny$\pm$0.16& 53.16\tiny$\pm$0.16 \\
            & Delta & 58.95\tiny$\pm$0.05 & 58.86\tiny$\pm$0.04 & 58.57\tiny$\pm$0.03 & 57.98\tiny$\pm$0.04 & 50.89\tiny$\pm$0.04 \\
            & BoT & \textbf{59.80}\tiny$\pm$0.07 & \textbf{59.77}\tiny$\pm$0.04 & \textbf{59.59}\tiny$\pm$0.03 & \textbf{59.04}\tiny$\pm$0.06 & \textbf{54.68}\tiny$\pm$0.03 \\
        \end{tabular}
        \caption{\textbf{Results on ImageNet-C.} BoT obtains better results than Tent and the 2 state-of-the-art methods in all cases. If the performance increase of BoT is not significant on ResNet50-BN (+0.78\% accuracy in average versus Delta), it is much more noticeable on ResNet50-GN (+4.31\% accuracy in average versus Delta) and VitBase-LN (+1.53.31\% accuracy in average versus Delta).}
        \label{tab:compmethods_imagenetc}                
    \end{table}
    
    \begin{table}[!tp]
        \centering
        \setlength\tabcolsep{4pt}
        \scriptsize
        \begin{tabular}{c|c|c|c|c|c|c}
            & \multirow{2}{*}{Method} & \multicolumn{5}{c}{Batch Size}\\\cline{3-7}
             & & 16 & 8 & 4 & 2 & 1 \\
            \hline
            \multirow{4}{*}{\tiny\rot{ResNet50-BN}} & Tent & 40.80\tiny$\pm$0.11 & 37.75\tiny$\pm$0.12 & 29.70\tiny$\pm$0.21 & 14.24\tiny$\pm$0.05 & 0.56\tiny$\pm$0.00 \\
            & SAR & 42.11\tiny$\pm$0.10 & 38.95\tiny$\pm$0.21 & 30.07\tiny$\pm$0.05 & 16.13\tiny$\pm$0.12 & 0.57\tiny$\pm$0.00 \\
            & Delta & 43.11\tiny$\pm$0.15 & 41.80\tiny$\pm$0.23 & 39.64\tiny$\pm$0.16 & 35.17\tiny$\pm$0.06 & 26.75\tiny$\pm$0.01 \\
            & BoT & \textbf{44.68}\tiny$\pm$0.24 & \textbf{43.12}\tiny$\pm$0.11 & \textbf{40.61}\tiny$\pm$0.22 & \textbf{35.55}\tiny$\pm$0.04 & \textbf{26.75}\tiny$\pm$0.00 \\
            \hline
            \multirow{4}{*}{\tiny\rot{ResNet50-GN}} & Tent & 39.35\tiny$\pm$0.16 & 39.29\tiny$\pm$0.18 & 39.28\tiny$\pm$0.19 & 39.27\tiny$\pm$0.18 & 39.26\tiny$\pm$0.18 \\
            & SAR & 42.94\tiny$\pm$0.108 & 42.75\tiny$\pm$0.05 & 42.28\tiny$\pm$0.09 & 41.75\tiny$\pm$0.06 & 41.84\tiny$\pm$0.05 \\
            & Delta & 43.10\tiny$\pm$0.05 & 43.11\tiny$\pm$0.05 & 42.74\tiny$\pm$0.12 & 41.89\tiny$\pm$0.10 & 42.18\tiny$\pm$0.03 \\
            & BoT & \textbf{44.21}\tiny$\pm$0.06 & \textbf{44.18}\tiny$\pm$0.10 & \textbf{43.84}\tiny$\pm$0.20 & \textbf{42.96}\tiny$\pm$0.16 & \textbf{42.49}\tiny$\pm$0.08 \\
            \hline
            \multirow{4}{*}{\tiny\rot{VitBase-LN}} & Tent & 43.28\tiny$\pm$1.04 & 42.81\tiny$\pm$1.04 & 42.48\tiny$\pm$0.87 & 42.28\tiny$\pm$1.05 & 42.49\tiny$\pm$1.32 \\
            & SAR & 52.72\tiny$\pm$0.19 & 52.59\tiny$\pm$0.25 & 52.20\tiny$\pm$0.16 & 50.92\tiny$\pm$0.11 & 49.95\tiny$\pm$0.18 \\
            & Delta & 53.32\tiny$\pm$0.23 & 53.31\tiny$\pm$0.28 & 53.03\tiny$\pm$0.24 & 52.25\tiny$\pm$0.34 & 49.76\tiny$\pm$0.20 \\
            & BoT & \textbf{54.63}\tiny$\pm$0.18 & \textbf{57.74}\tiny$\pm$0.19 & \textbf{54.62}\tiny$\pm$0.25 & \textbf{53.86}\tiny$\pm$0.28 & \textbf{51.91}\tiny$\pm$0.15 \\
        \end{tabular}
        \caption{\textbf{Results on ImageNet-Rendition.} The performance increase of BoT compared to Delta is similar on ResNet50-BN and ResNet50-GN (respectively +0.85\% and +0.87\% accuracy) but reaches +1.23\% accuracy on VitBase-LN.}
        \label{tab:compmethods_imagenetrendition}                
    \end{table}
    
    \begin{table}[!tp]
        \centering
        \setlength\tabcolsep{4pt}
        \scriptsize
        \begin{tabular}{c|c|c|c|c|c|c}
             & \multirow{2}{*}{Method} & \multicolumn{5}{c}{Batch Size}\\\cline{3-7}
             & & 16 & 8 & 4 & 2 & 1 \\
            \hline
            \multirow{4}{*}{\tiny\rot{ResNet50-BN}} & Tent & 27.82\tiny$\pm$0.30 & 22.47\tiny$\pm$0.40 & 10.71\tiny$\pm$0.42 & 2.94\tiny$\pm$0.08 & 0.13\tiny$\pm$0.00 \\
            & SAR & 31.05\tiny$\pm$0.29 & 26.73\tiny$\pm$0.20 & 16.80\tiny$\pm$0.07 & 6.72\tiny$\pm$0.05 & 0.13\tiny$\pm$0.00 \\
            & Delta & 31.92\tiny$\pm$0.11 & 30.36\tiny$\pm$0.16 & 27,32\tiny$\pm$0.16 & 22.56\tiny$\pm$0.16 & \textbf{15.58}\tiny$\pm$0.04 \\
            & BoT & \textbf{33.24}\tiny$\pm$0.13 & \textbf{31.50}\tiny$\pm$0.21 & \textbf{28.16}\tiny$\pm$0.12 & \textbf{22.86}\tiny$\pm$0.16 & \textbf{15.58}\tiny$\pm$0.04 \\
            \hline
            \multirow{4}{*}{\tiny\rot{ResNet50-GN}} & Tent & 23.04\tiny$\pm$0.40 & 22.95\tiny$\pm$0.38 & 22.93\tiny$\pm$0.38 & 22.92\tiny$\pm$0.38 & 22.92\tiny$\pm$0.35 \\
            & SAR & 32.11\tiny$\pm$0.50 & 32.26\tiny$\pm$0.07 & 31.89\tiny$\pm$0.16 & 31.16\tiny$\pm$0.20 & 31.64\tiny$\pm$0.25 \\
            & Delta & 34.50\tiny$\pm$0.20 & 34.26\tiny$\pm$0.09 & 33.57\tiny$\pm$0.18 & 31.56\tiny$\pm$0.08 & 30.93\tiny$\pm$0.07 \\
            & BoT & \textbf{35.77}\tiny$\pm$0.03 & \textbf{35.49}\tiny$\pm$0.19 & \textbf{34.91}\tiny$\pm$0.15 & \textbf{33.19}\tiny$\pm$0.10 & \textbf{32.07}\tiny$\pm$0.09 \\
            \hline
            \multirow{4}{*}{\tiny\rot{VitBase-LN}} & Tent & 5.83\tiny$\pm$0.32 & 5.69\tiny$\pm$0.43 & 5.59\tiny$\pm$0.44 & 5.38\tiny$\pm$0.28 & 5.51\tiny$\pm$0.49 \\
            & SAR & 25.40\tiny$\pm$0.65 & 25.88\tiny$\pm$0.64 & 27.87\tiny$\pm$0.08 & 32.89\tiny$\pm$0.57 & 30.68\tiny$\pm$0.99 \\
            & Delta & 38.67\tiny$\pm$0.08 & 38.50\tiny$\pm$0.08 & 38.18\tiny$\pm$0.11 & 37.18\tiny$\pm$0.14 & 33.90\tiny$\pm$0.08 \\
            & BoT & \textbf{39.69}\tiny$\pm$0.06 & \textbf{39.68}\tiny$\pm$0.06 & \textbf{39.50}\tiny$\pm$0.09 & \textbf{38.64}\tiny$\pm$0.03 & \textbf{34.09}\tiny$\pm$0.10 \\
        \end{tabular}
        \caption{\textbf{Results on ImageNet-Sketch.} BoT performs best in all case. The performance increase of BoT versus Delta is +0.72\% accuracy on ResNet50-BN, +1.32\% accuracy on ResNet50-GN and +1.03\% accuracy on VitBase-LN.}
        \label{tab:compmethods_imagenetsketch}                
    \end{table}

    \begin{table}[!tp]
        \centering
        \setlength\tabcolsep{4pt}
        \renewcommand*{\arraystretch}{1.1}
        \scriptsize
        \begin{tabular}{c|c|c|c|c|c|c}
             & \multirow{2}{*}{Method} & \multicolumn{5}{c}{Batch Size}\\\cline{3-7}
             & & 16 & 8 & 4 & 2 & 1 \\
            \hline
            \multirow{4}{*}{\rot{ResNet101}} & Tent & 65.30\tiny$\pm$0.08 & 64.65\tiny$\pm$0.18 & 63.47\tiny$\pm$0.12 & 58.89\tiny$\pm$0.33 & 49.10\tiny$\pm$0.04 \\
            & SAR & 63.08\tiny$\pm$0.03 & 57.47\tiny$\pm$0.05 & 46.20\tiny$\pm$0.09 & 24.81\tiny$\pm$0.16 & 18.63\tiny$\pm$0.01 \\
            & Delta & 73.20\tiny$\pm$0.08 & 71.52\tiny$\pm$0.12 & 68.16\tiny$\pm$0.11 & 61.41\tiny$\pm$0.20 & 49.08\tiny$\pm$0.03 \\
            & BoT & \textbf{73.54}\tiny$\pm$0.09 & \textbf{71.70}\tiny$\pm$0.07 & \textbf{68.17}\tiny$\pm$0.19 & \textbf{61.49}\tiny$\pm$0.10 & \textbf{50.28}\tiny$\pm$0.09 \\
        \end{tabular}
        \caption{\textbf{Results on VisDA2017.} Delta performs better than Tent and SAR but not as good as BoT. The performance improvement of BoT versus Delta is +0.36\% accuracy and +4.75\% versus Tent on ResNet101.}
        \label{tab:compmethods_visda2017}                
    \end{table}

\section{Conclusion}
    \label{sec:conclusion}
    In this work, we addressed the Fully Test-Time Adaptation problem when dealing with small batch sizes by analyzing the following tricks and methods: i) Usage of Batch renormalization or batch-agnostic normalization ii) Class re-balancing iii) Entropy-based sample selection iv) Temperature scaling.
    Our experimental results show that if those tricks used alone already yield an improved classification accuracy, using them in pairs is even better, and the best results are obtained by combining them all. By doing that, we significantly improve the current state-of-the-art across 4 different image datasets in terms of prediction performances. Furthermore, the selected tricks bring additional benefits concerning the computational load: i) Using group normalization instead of batch normalization in ResNet50 yields more stable results for the same number of total parameters ii) using the entropy-based sample selection improves the adapted model performance by using fewer samples. We hope that this study will be useful for the community and that the presented tricks and techniques will be integrated into future baselines and benchmarks.
    
\section{Acknowledgment}
    \label{sec:acknowledgment}
    This research was supported by the National Science and Engineering Research Council of Canada (NSERC), via its Discovery Grant program, and enabled in part by support provided by \href{https://www.calculquebec.ca/}{Calcul Québec} and the \href{https://alliancecan.ca}{Digital Research Alliance of Canada }.

    \clearpage
    {\small
    \bibliographystyle{ieee_fullname}
    \bibliography{egbib}
    }

\end{document}